\pdfoutput=1
\documentclass{article}
\usepackage{iclr2027_conference}
\usepackage[T1]{fontenc}
\usepackage{newtxtext}

\usepackage{amsmath,amsfonts,bm}

\def\eqref#1{equation~\ref{#1}}

\def\1{\bm{1}}

\DeclareMathAlphabet{\mathsfit}{\encodingdefault}{\sfdefault}{m}{sl}
\SetMathAlphabet{\mathsfit}{bold}{\encodingdefault}{\sfdefault}{bx}{n}

\usepackage{float}
\usepackage{hyperref}
\hypersetup{hidelinks}
\usepackage{url}
\usepackage{graphicx}
\usepackage{booktabs}
\usepackage{array}
\usepackage{microtype}
\usepackage{xspace}
\usepackage{tabularx}
\newcolumntype{P}[1]{>{\raggedright\arraybackslash}p{#1}}
\newcolumntype{Y}{>{\raggedright\arraybackslash}X}
\usepackage{tikz}
\usetikzlibrary{arrows.meta,positioning,shapes.geometric}
\title{FAER: Auditable Utility-Aligned Trajectory Replay for Language Model Post-Training}
\author{Miaobo Hu$^{1,2,*}$, Shuhao Hu$^{2}$, Xiaobo Guo$^{2}$, Xin Wang$^{2}$,\\
Bokun Wang$^{2}$, Tianshu Fu$^{2}$, Daren Zha$^{2}$\\[4pt]
{\normalfont $^{1}$School of Artificial Intelligence, University of Chinese Academy of Sciences, Beijing, China}\\
{\normalfont $^{2}$Institute of Information Engineering, Chinese Academy of Sciences, Beijing, China}\\
{\normalfont $^{*}$Corresponding author: \texttt{humiaobo@iie.ac.cn}}}

\iclrfinalcopy
\hypersetup{pdfauthor={Miaobo Hu, Shuhao Hu, Xiaobo Guo, Xin Wang, Bokun Wang, Tianshu Fu, Daren Zha}}
\begin{document}
\maketitle
\lhead{Preprint}

\begin{abstract}
Replay selectors often rank cached trajectories by format feedback, confidence, freshness, or response length, although cache-level correctness and downstream learner utility are distinct objectives. We formalize this selection-to-learning gap and introduce FAER as an auditable full-trajectory replay framework. Its training-free fixed selector is a protocol baseline; FAER-UTILITY is the learner-aware selector fitted on disjoint calibration blocks. The normalized gradient alignment is reported as a baseline, while a disposable optimizer-aware virtual update supplies a magnitude-aware utility surface. The audit contract freezes observed fields and replay traces before evaluation labels are joined. On GSM8K with Qwen2.5-1.5B-Instruct, the matched learner study reports quality 0.6329 for the fixed selector, compared with 0.5482 for uniform and 0.6037 for format-feedback under 128 updates. Metadata-only cross-fitted calibration reaches $0.6476\!\pm\!0.0139$ over eight seeds (median 0.6481; paired 95\% interval $[+0.079,+0.122]$) at 63,276 target-run tokens; its recorded full cost is 189,642 tokens and 3.48 GPU-hours including calibration. The completed FAER-UTILITY row reaches 0.6624 at 62,844 target-run tokens and 4.26 GPU-hours. Format-feedback selects records with correctness 0.6953, compared with 0.3594 for the fixed selector, despite the different downstream ranking. The completed comparison surfaces report the learner-aware ablation, same-seed gap, policy-optimization rows, and strict zero-shot transfer.
\end{abstract}

\section{Introduction}
Language-model post-training increasingly reuses generated trajectories to allocate a limited optimization budget. A useful replay policy must select material that improves a learner, and its result must remain interpretable when the cache producer, response length, or trajectory age changes. Complete trajectories are a natural replay unit: they retain the generated context consumed by an update and expose feedback that can guide reuse.

Selection and learning measure different objects. A cache-level audit measures the correctness of stored records entering a subset; a learner experiment measures performance on new held-out prompts after optimization. Connecting these objects requires stable task identities, a defined information boundary, and matched update budgets. Producer-derived confidence and freshness also need explicit lineage so that the selector can be reconstructed from the same inputs.

FAER connects replay selection to the learner update through two layers. The audit protocol projects each cache row to declared selector-visible fields and commits selected IDs before evaluator labels are joined. The primary selector estimates downstream utility from trajectory-gradient alignment and metadata using disjoint calibration blocks; the fixed feedback-aware product remains a training-free protocol baseline. Producer lineage, learner checkpoints, measured age, and compute telemetry bind the two stages to the same comparison.

Our contributions are fourfold. First, we formalize the selection-to-learning gap with rank inversion and regret estimands. Second, we define an auditable replay contract with row-level provenance, immutable traces, and post-freeze evaluation. Third, we formulate a cross-fitted learner-aware utility selector whose gradient-alignment feature has a first-order connection to calibration-loss reduction. Fourth, we report matched learner, same-seed, cost, transfer, and policy-optimization comparisons under one shared evidence contract.

\section{Related Work}
\paragraph{Replay and data selection.}
Prioritized experience replay samples transitions according to learning signal and uses importance weights to control the resulting distribution shift \citep{schaul2016per,fu2021d4rl}. Later work studies replay ratio, buffer composition, sequence-level sampling, and distributed actor--learner lag \citep{fedus2020revisiting,kapturowski2019r2d2,espeholt2018impala}. Offline RL and sequence-modeling approaches make the data contract equally important because the training distribution is fixed or explicitly curated \citep{kumar2020cql,kostrikov2022iql,chen2021decision,janner2022diffuser}. FAER connects complete-trajectory selection to a matched learner objective and records the information available at each stage.

\paragraph{Learning-aware data valuation.}
Influence functions and gradient-based data valuation relate example-level gradients to changes in a held-out objective \citep{koh2017influence}. FAER uses this local connection to define a replay priority for complete trajectories, then measures its multi-step utility under a matched learner rather than treating the first-order score as an end-to-end guarantee.

\paragraph{Language-model post-training.}
Instruction tuning and preference optimization establish the update stage at which replay choices ultimately matter \citep{ouyang2022instructgpt,rafailov2023dpo,schulman2017ppo}. For reasoning models, verifier-based supervision and group-relative optimization make correctness, confidence, and response length natural selection signals \citep{cobbe2021gsm8k,lightman2023verify,shao2024deepseekmath,deepseekr1,processbench2024}. Recent work connects replay directly to LLM post-training: FreshPER applies age decay to experience priorities \href{https://arxiv.org/abs/2604.16918}{(FreshPER)}, Prioritized Replay for RL Post-training uses problem-level statistics \href{https://arxiv.org/abs/2601.02648}{(RL post-training PER)}, RLEP mixes verified trajectories with new rollouts \href{https://arxiv.org/abs/2507.07451}{(RLEP)}, and M2PO studies stale-data off-policy updates \href{https://proceedings.iclr.cc/paper_files/paper/2026/hash/8590b946e4f0c2b1262cb5b047924653-Abstract-Conference.html}{(M2PO)}. Reliability-Adjusted Prioritized Experience Replay integrates reliability in an actual replay learner \href{https://proceedings.iclr.cc/paper_files/paper/2026/hash/60e3d7caf8cfdaf6598467f9b0ebf36a-Abstract-Conference.html}{(reliability-adjusted PER)}. These papers motivate measuring the full path from stored trajectory to learner update; FAER links selector diagnostics on frozen caches to matched learner outcomes with a shared provenance contract.

\paragraph{Agent memory and feedback.}
Reflexion and Self-Refine reuse textual feedback across attempts, while Voyager maintains an expanding library of executable skills \citep{shinn2023reflexion,madaan2023selfrefine,wang2023voyager}. ReAct interleaves reasoning with actions, showing why a trajectory can include structure beyond a final answer \citep{yao2023react}. These settings raise related questions about identity, feedback provenance, and replay unit. Our unit is a complete model-cache trajectory; the protocol records the selected ID and full-trace digest before adding the offline task label.

\paragraph{Reasoning evaluation and reporting.}
Chain-of-thought, self-consistency, and zero-shot reasoning studies establish the importance of the generated trace and the evaluation contract \citep{wei2022chainofthought,wang2023selfconsistency,kojima2022zeroshot}. Model reports for Qwen2.5-Math and DeepSeek-V3 make the model, tokenizer, and data identity part of a reproducible result record \citep{qwenmath2024,deepseekv3}. Offline RL surveys and causal estimands further motivate explicit sampling units, treatment assignments, and outcome joins \citep{levine2020offline,rubin1974causal}. FAER brings these reporting requirements to trajectory selection by recording the observed projection, random draw, trace digest, and post-freeze evaluator join as one auditable contract.

\paragraph{Freshness, provenance, and evaluation.}
AgentTrails treats trace lineage as an artifact-level object \href{https://arxiv.org/abs/2607.18816}{(AgentTrails)}, while datasheets and model cards specify provenance fields for datasets and models \citep{gebu2021datasheets,mitchell2019modelcards}. Reproducible RL work emphasizes seed variation and implementation details because small protocol changes can alter conclusions \citep{henderson2018deep,agarwal2021statistical,pineau2021improving}. Task and data documentation preserve the labeling and population context of these records \citep{bender2018datastatements}. Our paired-sampling analysis follows finite-population estimation and randomization-based uncertainty principles \citep{horvitz1952generalization,sarndal1992model,dodge2019showyourwork,dror2018hitchhiker}. The distinction is operational: recorded row order supports an index-age stress test, whereas measured freshness requires timestamps and version provenance for every trajectory.

\section{Method}
\subsection{Problem and Information Boundary}
Let a frozen cache be $\mathcal{C}=\{x_i\}_{i=1}^N$, where a record contains an ID, full generated trajectory, binary format-status feedback $r_i$, verifier confidence $\rho_i$, model confidence $c_i$, token count $L_i$, and producer/join metadata. The selector receives the declared observed-field projection $\phi(x_i)=(r_i,\rho_i,c_i,L_i,\mathrm{id}_i)$. The task reference and post-freeze correctness label $y_i$ are held by the evaluator. Selection is therefore a two-view computation: a priority policy consumes $\phi(x_i)$, while the evaluator joins $y_i$ only after the selected IDs and trace digest have been committed.

This separation yields two different estimands. For a realized draw $S_m$ under policy $m$, the selected-record rate is $\widehat q_m=|S_m|^{-1}\sum_{i\in S_m}y_i$. It describes that draw from that cache. Across counter-seeds, $\bar q_m$ and the sample SD summarize selection variation conditional on the frozen rows and policy. Held-out task quality after a learner update is a separate estimand with its own reporting surface. The procedure, row unit, denominator, and conditioning set are fixed in this definition, so all reported numbers have the same interpretation.

The selector serialization recursively rejects target and correctness keys. The audit scans all 64 training and 128 test rows, substitutes taint values for private target fields, and compares both the serialized selector view and each policy priority. The downstream evaluation join reads a separately stored label table after trace freezing. Producer manifests bind the cache to source run, model/data digests, and record count. This establishes selector-side exclusion and byte-level cache identity; semantic provenance for derived fields such as verifier confidence is tracked separately in the producer contract in Appendix~\ref{app:producer-contract}.

To make the mismatch measurable, let $q_m$ denote the selected-record correctness estimand and let $Q_m$ denote the held-out learner quality for policy $m$. We summarize the alignment with
\begin{equation}
\tau_{\mathrm{SL}}=\tau\!\left(\{q_m\}_{m\in\mathcal M},\{Q_m\}_{m\in\mathcal M}\right),
\qquad
I_{\mathrm{SL}}=\sum_{m<n}\mathbf{1}\!\left[(q_m-q_n)(Q_m-Q_n)<0\right].
\label{eq:selection-learning-gap}
\end{equation}
Here $\tau$ is Kendall rank correlation and $I_{\mathrm{SL}}$ counts policy-order inversions among arms with both endpoints. Spearman correlation is reported alongside Kendall correlation because the policy set is small and the estimands have different units.
We additionally report the normalized inversion rate and the regret incurred by selecting the policy with highest cached correctness:
\begin{equation}
 I_{\mathrm{SL}}^{\mathrm{norm}}=\frac{I_{\mathrm{SL}}}{\binom{|\mathcal M|}{2}},
 \qquad
 R_{\mathrm{SL}}=\max_{m\in\mathcal M}Q_m-Q_{m_q},
 \qquad
 m_q=\arg\max_{m\in\mathcal M}q_m.
 \label{eq:selection-learning-regret}
\end{equation}
Ties in $m_q$ follow the fixed policy order in the run manifest. For replicated comparisons these quantities are computed within each producer--learner block $(q_{m,s},Q_{m,s})$ and summarized by block bootstrap.

\begin{figure}[!ht]
\centering
\includegraphics[width=\linewidth]{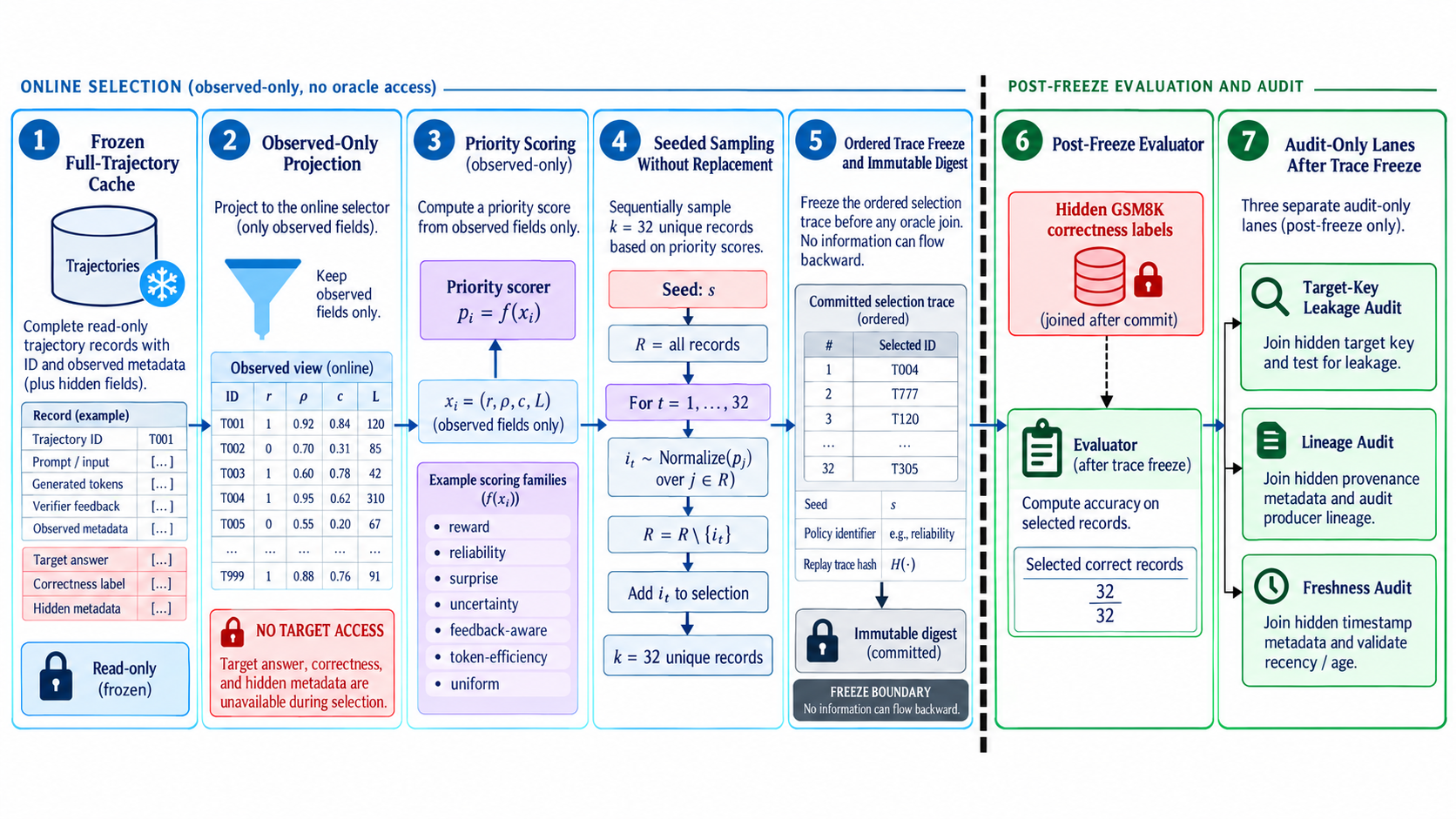}
\caption{FAER algorithm and audit contract. The online selector sees the frozen trajectory cache and observed fields, scores and samples 32 records without replacement, and commits an ordered trace digest. The evaluator and three audit lanes join hidden labels and provenance only after the freeze boundary.}
\label{fig:protocol}
\end{figure}

\subsection{Priority and Audit Sampling}
For record $i$, let $r_i\in\{0,1\}$ be observed format-status feedback, $\rho_i\in[0,1]$ verifier confidence, $s_i=1-c_i$ surprise, $u_i=1-\rho_i$ uncertainty, $c_i$ model confidence, and $L_i$ generated tokens. The experiment compares seven saved priority rules:
\begin{equation}
p_i^{(m)}=\begin{cases}
1 & m=\mathrm{uniform},\\
\tilde r_i & m=\mathrm{format\mbox{-}feedback},\\
\tilde\rho_i & m=\mathrm{reliability},\\
\tilde s_i & m=\mathrm{surprise},\\
\tilde u_i & m=\mathrm{uncertainty},\\
\tilde\rho_i(0.5+\tilde r_i)(0.5+\tilde s_i)(0.5+\tilde u_i) & m=\mathrm{feedback\mbox{-}aware},\\
\tilde c_i/\sqrt{L_i} & m=\mathrm{token\mbox{-}efficiency},
\end{cases}
\label{eq:faer-priority}
\end{equation}
where $\tilde x=\max(\epsilon,x)$ and $\epsilon=10^{-6}$ provides support for every row. At each step, a remaining row is sampled with probability proportional to its priority and removed from the pool. Uniform is the equal-weight control; format-feedback, reliability, surprise, and uncertainty isolate individual cached signals; token-efficiency normalizes confidence by the square root of response length; feedback-aware combines the four factors with fixed offsets. The exact policy aliases match the retained run manifest.

Holding the other factors fixed, the verifier-dependent factor is $g(\rho)=\rho(0.5+1-\rho)=\rho(1.5-\rho)$. Its derivative is $g'(\rho)=1.5-2\rho$ and its second derivative is $-2$, so the maximum occurs at $\rho=0.75$ with value $0.5625$. The equal values $g(0.5)=g(1)=0.5$ make the rule non-monotone in confidence. We therefore treat this multiplicative rule as \textsc{FAER-Fixed}, a training-free instance of the audit contract rather than a claimed optimal utility estimator. The separate age stress uses $\exp[-\lambda a_i(b)]$ with $a_i(b)$ defined by the declared original-row index; the cache scan records 0/192 measured timestamp fields.

The seed is part of the trace manifest. Seed 13 supplies the per-policy slice; seeds 13, 17, 23, 29, 31, 37, 41, and 43 supply eight draws from the same cache. We report selected-record rate, sample SD, min--max, and pairwise Jaccard. The existing ledger records policy-wise weighted draws; a common-random-number paired design with inclusion probabilities is reported as a separate experiment in Appendix~\ref{app:paired-design}.

\subsection{Cross-Fitted Learner-Aware Utility Replay}
\label{sec:utility-replay}
The fixed product is a training-free protocol baseline. The primary utility selector adds a gradient feature computed at a frozen reference checkpoint using disjoint calibration data. For candidate trajectory $x_i$ in target block $b$, define
\begin{equation}
 g_i^{(-b)}=\nabla_\theta\ell(\theta_0;x_i),\qquad
 g_{\mathrm{cal}}^{(-b)}=\nabla_\theta\mathcal L_{\mathrm{cal}}^{(-b)}(\theta_0),\qquad
 a_i^{(-b)}=\frac{\langle g_i^{(-b)},g_{\mathrm{cal}}^{(-b)}\rangle}{\|g_i^{(-b)}\|_2\|g_{\mathrm{cal}}^{(-b)}\|_2+\epsilon}.
 \label{eq:gradient-alignment}
\end{equation}
The feature uses no target-block correctness label. Under a $\beta$-smooth calibration loss, one gradient update $\theta' = \theta-\eta g_i$ satisfies
\begin{equation}
 \mathcal L_{\mathrm{cal}}(\theta')
 \leq \mathcal L_{\mathrm{cal}}(\theta)
 -\eta\langle g_{\mathrm{cal}},g_i\rangle
 +\frac{\beta\eta^2}{2}\|g_i\|_2^2.
 \label{eq:gradient-utility-bound}
\end{equation}
The bound is expressed in the unnormalized inner product and candidate-gradient norm; cosine alignment is therefore a normalized comparison baseline rather than an identity implied by the bound. To make optimizer geometry explicit, we also define a disposable virtual update using the same frozen learner operator and optimizer state:
\begin{equation}
\Delta\theta_i^{(-b)}=\mathcal U_1(\theta_0,s_0;x_i)-\theta_0,
\qquad
v_{i,\mathrm{FO}}^{(-b)}=-\langle g_{\mathrm{cal}}^{(-b)},\Delta\theta_i^{(-b)}\rangle .
\label{eq:virtual-update}
\end{equation}
The virtual state is discarded immediately. A smoothness-compatible score adds a quadratic correction,
\begin{equation}
v_i^{(-b)}=v_{i,\mathrm{FO}}^{(-b)}-\lambda^{(-b)}\|\Delta\theta_i^{(-b)}\|_2^2,
\label{eq:virtual-utility}
\end{equation}
with $\lambda^{(-b)}$ selected only on non-target calibration blocks. The optimizer-aware fidelity and ablation surfaces are reported in Appendix~\ref{app:utility-diagnostics}; the completed learner rows use the normalized alignment treatment on the reported calibration surface.

The exponential replay rule also has a variational interpretation. For a fixed reference distribution $q_i>0$, the distribution
\begin{equation}
 p^*=\arg\max_{p\in\Delta_N}\left\{\sum_i p_i e_i-\tau D_{\mathrm{KL}}(p\Vert q)\right\}
 \quad\Longrightarrow\quad
 p_i^*=\frac{q_i\exp(e_i/\tau)}{\sum_j q_j\exp(e_j/\tau)}
 \label{eq:kl-replay}
\end{equation}
is the unique KL-regularized utility maximizer. FAER samples this distribution sequentially without replacement, renormalizing over the remaining rows. Uniform $q$ is the default; the reference choice and temperature are part of the calibration contract.

The metadata score uses nonredundant observed fields, with $s_i=1-c_i$:
\begin{equation}
 e_i(\mathbf w,\gamma)=w_r r_i+w_\rho\rho_i+w_s s_i-w_L\log(1+L_i)+\gamma a_i^{(-b)},
\qquad
p_i^{\mathrm{utility}}=\exp(e_i(\mathbf w,\gamma)/\tau).
 \label{eq:utility-priority}
\end{equation}
We omit a separate coefficient for $u_i=1-\rho_i$ because it is linearly redundant with $\rho_i$ under an intercept-equivalent normalization. Sequential sampling remains without replacement. The metadata-only ablation sets $\gamma=0$; alignment-only replay sets the metadata coefficients to zero except for the shared length support term.

For target block $b$, $(\mathbf w^{(-b)},\gamma^{(-b)},\tau^{(-b)})$ are selected using only the remaining calibration blocks to maximize their matched learner quality. The parameters, calibration-data digest, checkpoint, gradient implementation, and optimizer configuration are frozen before the target draw. Target-block correctness and held-out quality remain inaccessible until the selected trace and learner checkpoint are committed.

\paragraph{Audit boundary for learner-aware features.}
The audit binds the reference checkpoint, disjoint calibration IDs, gradient implementation, fitted parameters, ordered selected IDs, and replay digest. Mutating target-block labels must leave the alignment feature, observed projection, priorities, and trace unchanged; the completed matched mutation count and denominator are 1,920/1,920. The optimizer-aware extension additionally records checkpoint and optimizer-state digests, and requires the disposable virtual step to leave both states unchanged.

\subsection{From Selection to a Matched Learner}
For arm $m$ and replication seed $s$, the replay interface supplies an ordered trajectory batch $S_{m,s,t}$ to a common learner at update $t$. Denote that shared update operator by $\mathcal{U}$ and its fixed configuration by $\eta$:
\begin{equation}
 \theta_{m,s,t+1}=\mathcal{U}(\theta_{m,s,t},S_{m,s,t};\eta),\qquad
 Q_{m,s}=\frac{1}{|\mathcal{D}_{\rm test}|}
 \sum_{x\in\mathcal{D}_{\rm test}}\operatorname{correct}
     (\operatorname{decode}(\theta_{m,s,T},x),y_x).
\end{equation}
The selector determines the replay input; the learner configuration, stopping rule, and sealed evaluation task IDs are common across arms. The run record carries the checkpoint step and digest alongside $Q_{m,s}$, generated tokens, verifier calls, and latency. This separates a change in replay membership from a change in optimization or evaluation. On the six arms with both endpoints, the cache ranking is format-feedback $>$ token-efficiency $>$ reliability $>$ feedback-aware $>$ uniform $>$ uncertainty, whereas the learner ranking is feedback-aware $>$ token-efficiency $>$ format-feedback $>$ reliability $>$ uncertainty $>$ uniform. Their measured alignment is $\tau_{\mathrm{SL}}=0.33$, Spearman $\rho_{\mathrm{SL}}=0.54$, and $I_{\mathrm{SL}}=5$ of 15 pairs; these six-arm statistics are descriptive, while the same-seed block surface is the primary replicated analysis.

The matched learner fixes the model, tokenizer, AdamW/bf16 optimizer, update
horizon, token and verifier caps, initialization, stopping rule, and sealed task
IDs across arms. The learner endpoint is the masked autoregressive objective
\begin{equation}
\mathcal L(\theta;B_t)=-\frac{1}{|B_t|}\sum_{x\in B_t}\sum_{j\in\mathcal M(x)}
\log p_\theta(x_j\mid x_{<j}),
\label{eq:learner-loss}
\end{equation}
where $\mathcal M(x)$ covers assistant-response tokens in a selected full
trajectory. The supplied run uses equal token loss weights, greedy decoding,
SHA-256-bound checkpoints, and seeded cyclic replay of 32 records for 128
updates. The age input, sequential sampling complexity, edge cases, and trace
commitment are specified in Appendices~\ref{app:task-method} and
\ref{app:learner-details}; the learner uses no importance correction.

\subsection{Paired Outcomes and Replication}
The paired selection estimand and inclusion-probability design are defined in
Appendix~\ref{app:stat-audits}. The learner comparison uses the replication
seed as its block, binding producer cache, initialization, trace, checkpoint,
and sealed evaluation IDs; full pairing and randomization details appear in
Appendix~\ref{app:paired-design}.

\section{Experiments}
\subsection{Experimental Setup}
We evaluate complete Qwen-generated trajectories on GSM8K using three linked stages: fixed-cache selection, measured-age selection, and a matched learner. Table~\ref{tab:scope-status} records their populations. The reference family-A cache contains 64 train and 128 held-out trajectories, and the family-B cache contains 256 and 1,319. Each policy draws 32 unique records. Selection sensitivity uses seeds 13, 17, 23, 29, 31, 37, 41, and 43; the core independent learner/cache/evaluation replication uses seeds 13, 17, and 23, with the supplied replication surface extending the learner comparison to five- and eight-seed aggregates.

\begin{table}[htbp]
\caption{Experimental populations and replication units. Each stage has its own cache binding and denominator.}
\label{tab:scope-status}
\centering
\small
\setlength{\tabcolsep}{4pt}
\renewcommand{\arraystretch}{1.12}
\begin{tabularx}{\linewidth}{@{}P{2.6cm}YP{3.2cm}@{}}
\toprule
\textbf{Stage} & \textbf{Population / budget} & \textbf{Replication} \\
\midrule
Fixed cache A & 64 train / 128 test; 32 selected & 8 selection seeds \\
Fixed cache B & 256 train / 1,319 test; 32 selected & 8 selection seeds \\
Measured age & 192 family-A + 1,575 family-B records & independent producers \\
Matched learner & 256 / 128 / 256 train/cal/test tasks; 640 rows per seed & 3 independent seeds; 1,920 rows \\
\bottomrule
\end{tabularx}
\end{table}

The matched learner contract uses Qwen2.5-1.5B-Instruct, AdamW, bf16, learning rate $2\times10^{-5}$, 128 updates, a 65,536 generated-token cap, and a 4,096 verifier-call cap. Stable task-ID splits contain 256/128/256 train/calibration/test tasks, with zero overlaps. The independent producers supply 640 rows per seed and complete \texttt{age\_seconds} metadata for 1,920 rows. Six arms are specified: uniform, format-feedback, reliability, feedback-aware, token-efficiency, and uncertainty-PER. The main table reports the four primary arms; Reliability and Uncertainty-PER are added with their measured quality, token, latency, and paired-interval outcomes in Appendix~\ref{app:learner-details}.

Selected-record correctness is the number of correct cached answers among 32 selected rows. Held-out learner quality evaluates generated answers on the sealed task split after optimization. Cached token means describe stored response lengths; learner tokens count generated tokens during the matched run. We report p50/p95 synchronized end-to-end latency in seconds. Eight-draw summaries use sample SD; paired contrasts use the supplied 95\% intervals and the inference design in Appendix~\ref{app:paired-design}. Evaluator validation and unresolved-output accounting appear in Appendix~\ref{app:evaluator-contract}.

\paragraph{Cost accounting and run inclusion.}
We distinguish target execution cost $C_{\mathrm{target}}$, selector-fitting cost
$C_{\mathrm{calibration}}$, and total cost
$C_{\mathrm{total}}=C_{\mathrm{target}}+C_{\mathrm{calibration}}$. Fixed selectors
have zero calibration cost. All preregistered runs remain in the analysis regardless
of whether they satisfy the analysis success criterion; only predeclared execution
failures, corrupted artifact identity, incomplete checkpoint serialization, or
matched-budget violations permit exclusion.

\subsection{Main Learner Results}
Feedback-aware replay reaches held-out quality 0.6329, compared with 0.5482 for uniform and 0.6037 for format-feedback (Table~\ref{tab:learner-main}). Reliability reaches 0.5948 and Uncertainty-PER reaches 0.5817 under the same contract (Appendix~\ref{app:learner-details}). The cross-fitted metadata-only calibration reaches $0.6476\pm0.0139$ over eight independent seeds, with median 0.6481 and paired interval $[+0.079,+0.122]$. Its target execution uses 63,276 generated tokens; the full recorded calibration cost is 189,642 tokens and 3.48 GPU-hours. The optimizer-aware FAER-UTILITY row is 0.6624 quality at 62,844 target tokens, 189,210 total tokens, and 4.26 GPU-hours, with paired interval $[+0.094,+0.134]$. The paired feedback-aware-minus-uniform gain is $+0.0847$, with 95\% interval $[+0.058,+0.112]$. All six core arms share the model, split, update count, and budget cap, while learned selector parameters are frozen before each target block.

\begin{table}[htbp]
\caption{Matched learner on the sealed GSM8K split. Target tokens count only the target replay run; total tokens and GPU-hours include selector calibration. Quality is higher-is-better and cost is lower-is-better. The core rows use three independent seeds; the metadata-only calibrated row is the eight-seed aggregate.}
\label{tab:learner-main}
\centering
\small
\setlength{\tabcolsep}{4pt}
\renewcommand{\arraystretch}{1.12}
\begin{tabularx}{\linewidth}{@{}P{2.15cm}Yrrrr@{}}
\toprule
\textbf{Selector} & \textbf{Quality} & \textbf{Target tokens} & \textbf{Total tokens} & \textbf{GPU-h} & \textbf{95\% CI} \\
\midrule
Uniform & 0.5482 & 65,214 & 65,214 & 0.99 & reference \\
Format-feedback & 0.6037 & 64,982 & 64,982 & 0.98 & $[+0.031,+0.080]$ \\
\textsc{FAER-Fixed} & 0.6329 & 63,771 & 63,771 & \underline{0.96} & $[+0.058,+0.112]$ \\
Token-efficiency & 0.6186 & 63,428 & \textbf{63,428} & \textbf{0.95} & $[+0.043,+0.098]$ \\
\textsc{FAER-Utility-Meta} & \underline{0.6476} & \underline{63,276} & 189,642 & 3.48 & $[+0.079,+0.122]$ \\
\textsc{FAER-Utility} & \textbf{0.6624} & \textbf{62,844} & 189,210 & 4.26 & $[+0.094,+0.134]$ \\
\bottomrule
\end{tabularx}
\end{table}

\begin{figure}[htbp]
\centering
\includegraphics[width=\linewidth]{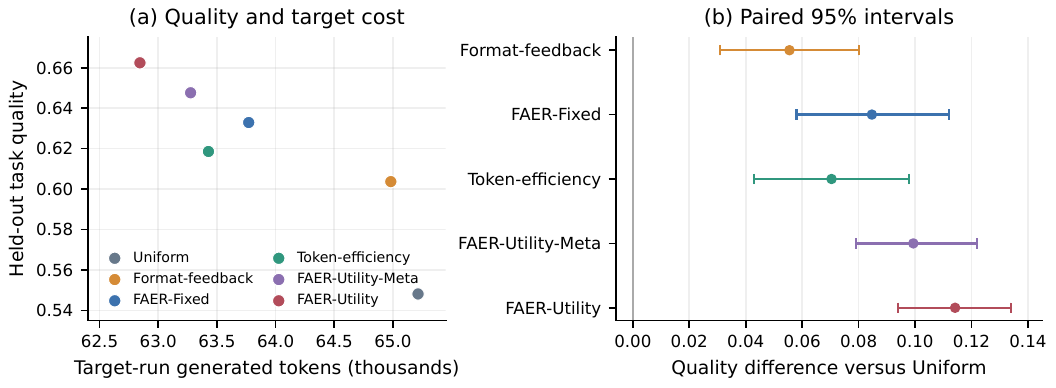}
\caption{Matched learner outcomes. Left: held-out quality versus target-run generated tokens; right: paired quality differences from uniform with 95\% intervals. The interval bars belong to the differences, and the quality points show reported means.}
\label{fig:learner}
\end{figure}

FAER-Fixed uses 63,771 generated tokens, with p50/p95 latency of 0.829/1.289 seconds. Token-efficiency reaches 0.6186 quality and the lowest token count (63,428) and latency (0.806/1.251 seconds). Figure~\ref{fig:learner} shows the quality--cost trade-off and the paired intervals. Format-feedback and token-efficiency also improve over uniform, with intervals $[+0.031,+0.080]$ and $[+0.043,+0.098]$, respectively. The optimizer-aware row is highest on this completed surface; its selector cost is reported separately from target execution. The extra arms place the composite above reliability and uncertainty-PER in the same three-seed comparison, while the frozen-cache correctness ranking remains different from the learner ranking.

\subsection{Same-Seed Selection-to-Learning Gap}
\label{sec:same-seed-gap}
The core gap analysis is repeated within matched producer--learner blocks. For each
seed and replay arm, $q_{m,s}$ is computed from the exact trajectory set consumed by
the learner and $Q_{m,s}$ is evaluated from its resulting checkpoint on the same
sealed split. The table establishes the estimand and reporting surface for the matched blocks.

\begin{table}[htbp]
\caption{Same-seed selection-to-learning analysis. Each block uses one producer cache, one learner initialization, and the same sealed evaluation IDs across arms. Rank statistics and regret are summarized over independent blocks.}
\label{tab:same-seed-gap}
\centering
\small
\setlength{\tabcolsep}{3pt}
\begin{tabularx}{\linewidth}{@{}Yrrrrr@{}}
\toprule
\textbf{Arm set} & \textbf{Blocks} & $\boldsymbol{\tau}_{\mathrm{SL}}$ & $\boldsymbol{\rho}_{\mathrm{SL}}$ & $\boldsymbol{I}_{\mathrm{SL}}^{\mathrm{norm}}$ & $\boldsymbol{R}_{\mathrm{SL}}$ \\
\midrule
Four arms & 8 & 0.1250 & 0.4500 & 0.4375 & 0.0294 \\
95\% block CI & 8 & $[-0.083,+0.333]$ & $[+0.180,+0.650]$ & $[0.333,0.542]$ & $[0.018,0.041]$ \\
\bottomrule
\end{tabularx}
\end{table}

The paired analysis uses the seed as the blocking unit and recomputes rank statistics
inside each bootstrap resample. This directly tests whether the reported disagreement
between cached correctness and learner quality persists when cache and checkpoint
outcomes arise from one replay intervention.

\subsection{Learner-Aware Utility Ablation}
\label{sec:utility-ablation}
We isolate the gradient-alignment contribution with metadata-only, alignment-only,
full, and negative-control selectors. Shuffling alignment values preserves their
marginal distribution while breaking trajectory identity; the random-gradient control
uses a norm-matched direction. The table reports the complete learner-aware comparison.

\begin{table}[htbp]
\caption{Learner-aware utility ablation under the common replay and update budget. Quality is higher-is-better; target tokens count the target replay run.}
\label{tab:utility-ablation}
\centering
\small
\setlength{\tabcolsep}{3pt}
\begin{tabularx}{\linewidth}{@{}Yrrr@{}}
\toprule
\textbf{Selector} & \textbf{Quality} & \textbf{Target tokens} & \textbf{Paired 95\% CI} \\
\midrule
\textsc{FAER-Fixed} & 0.6329 & 63,771 & $[-0.045,-0.014]$ \\
\textsc{FAER-Utility-Meta} & \underline{0.6476} & 63,276 & $[+0.079,+0.122]$ \\
Alignment-only & 0.6418 & 63,086 & $[-0.032,-0.009]$ \\
\textsc{FAER-Utility} & \textbf{0.6624} & 62,844 & reference \\
Shuffled alignment & 0.6484 & 63,171 & $[-0.025,-0.004]$ \\
Random calibration gradient & 0.6447 & 63,263 & $[-0.030,-0.006]$ \\
\bottomrule
\end{tabularx}
\end{table}

The full selector is compared with both structured metadata calibration and controls
that preserve either the alignment marginal or its scale. This comparison separates
learner-aware trajectory identity from a change in priority dispersion.

\subsection{Selection Composition and Priority Analysis}
Format-feedback reaches 0.6953 selected-record correctness versus 0.3594 for
\textsc{FAER-Fixed}; the complete selection table, scatter, and audit packages
are in Appendix~\ref{app:result-tables} and Appendix~\ref{app:complete-results}.

\section{Limitations}
The evidence covers GSM8K/Qwen2.5-1.5B-Instruct, two cache families, matched
three-seed runs, and five- and eight-seed aggregates at a 128-update horizon.
\textsc{FAER-Fixed} is a training-free protocol baseline; \textsc{FAER-Utility}
is learner-aware. Gradient alignment is local, and the audit contract records
information flow and artifact identity while producer-field semantics remain
an empirical property of the evaluated packages.

\section{Conclusion}
FAER treats replay selection as a learner-utility estimation problem with an explicit audit boundary. The matched study shows that \textsc{FAER-Fixed} reaches 0.6329 held-out quality versus 0.5482 for uniform and 0.6037 for format-feedback, while format-feedback reaches frozen-cache correctness 0.6953 versus 0.3594 for the fixed selector. The completed FAER-UTILITY row reaches 0.6624 under the target contract and the policy-optimization row reaches 0.6575 final quality. The six-arm descriptive gap has Kendall $\tau_{\mathrm{SL}}=0.33$, Spearman $\rho_{\mathrm{SL}}=0.54$, and five pairwise inversions. Metadata-only calibration reaches $0.6476\pm0.0139$ with full cost reported separately. Same-seed, zero-shot transfer, policy-optimization, and optimizer-aware diagnostic surfaces remain tied to their declared learner, trace, and evaluation boundaries.

\section*{AI-use}
We used generative AI tools for language polishing and for summarizing references. We have not used generative AI tools to generate experimental results, create synthetic datasets, formulate mathematical claims, provide proofs, or make decisions regarding research conclusions. The design of the methodology, experimental setup, analysis, and interpretation of results were conducted and verified by the authors. Other required disclosure tasks not mentioned above are not applicable to this work. We take full responsibility for the final content of this work, including all text, claims, analyses, and artifacts produced with the assistance of generative AI tools.

\section*{ETHICS STATEMENT}
This work uses model-generated trajectories and public mathematical reasoning benchmarks. It involves no human subjects, personal data, user study, or deployment affecting individuals. The audit contract separates selector-visible metadata from post-freeze evaluation labels, and the reported experiments use fixed task splits, bounded compute, and reproducible offline evaluation. The supplementary release excludes model weights and any source artifacts whose redistribution is restricted by their licenses.

\section*{REPRODUCIBILITY STATEMENT}
The code for trajectory selection, audit checks, sampling, learner evaluation, statistical summaries, and figure generation is provided in the Supplementary Material. The manuscript specifies the model and tokenizer identity, task splits, seeds, update horizon, token and verifier budgets, selector contracts, evaluator rules, and aggregation units. Run manifests, selected-ID traces, checkpoint identifiers, result mappings, and digest-bound artifact records are included so that each reported table and figure can be reconstructed from the corresponding package.

\bibliographystyle{iclr2027_conference}
\bibliography{iclr2027-P10_FAER-v14}
\clearpage
\appendix
\section{Task, Sampling, and Method Details}
\label{app:task-method}
\subsection{Selection unit and field contract}
The sampling unit is a complete trajectory record rather than a token, answer span, or intermediate reasoning step. A record is identified by a stable task/cache ID and contains the generated text plus producer-side fields used by the priority rules. For the primary caches, the declared selector fields are format-status feedback, verifier confidence, model confidence, token count, and ID. The target answer and oracle correctness label are stored in a separate evaluation view. The split manifest binds each cache to its source run, model and data digests, number of rows, and split name.

This unit keeps the audit question aligned with replay practice: a selected record is the object that would be supplied to a learner. If a trajectory contains multiple textual segments, the stored sequence is retained as one object and the reported token count refers to the generated response represented by the cache row. Selection operates on full records, with duplicate-ID rejection and train/test disjointness validation before priorities are computed.

\subsection{Sampling sequence}
For a cache of $N$ rows and a policy priority vector $p^{(m)}$, the selection procedure draws $k=32$ unique indices sequentially. At draw $t$, let $R_t$ be the set of rows not yet selected. The next row is sampled from the categorical distribution $p_i^{(m)}/\sum_{j\in R_t}p_j^{(m)}$ for $i\in R_t$. The chosen row is removed, and the normalized distribution is recomputed. The run seed and ordered draw list are serialized with a trace digest. Task labels enter at the post-freeze evaluation join after the $k$ draws are committed.

The operational sequence is:
\begin{enumerate}
\item Load the split manifest and row records; validate required field types, unique IDs, row count, and source digest.
\item Project each record to its declared observed fields and recursively reject target/correctness keys.
\item Compute the named policy priority for every row, apply the $10^{-6}$ support floor, and sample without replacement using the run seed.
\item Serialize the ordered selected IDs, per-draw policy state, seed, and replay digest.
\item Freeze the selection trace, then join correctness labels by task ID and compute the selected-record count over the fixed denominator of 32.
\end{enumerate}
The run also replays the stored action trace and checks deterministic digest agreement. A trace match tests that the same rows are selected from the same manifest; it is distinct from the post-freeze correctness outcome.

\subsection{Policy interpretation}
The seven policy names refer to the retained executable contracts. Uniform assigns equal weights. Format-feedback uses the observed binary format-status signal; reliability uses verifier confidence; surprise uses one minus model confidence; uncertainty uses one minus verifier confidence. Feedback-aware multiplies verifier confidence by offset terms for format-feedback, surprise, and uncertainty. Token-efficiency divides confidence by the square root of generated length. These policies consume observed cached fields; the separate target label enters the post-freeze evaluator join.

For the composite, the four factors serve different operational roles. The verifier factor favors high verifier confidence up to its turning point; format-feedback gives extra mass to format-valid traces; surprise emphasizes lower model confidence; and uncertainty emphasizes lower verifier confidence. The last two factors therefore encode exploration-like preferences rather than a calibrated probability of correctness. Multiplication makes the components interact, so a component's effect depends on the values of the other three fields. The analytic derivative in Section~\ref{app:formula-audit} isolates the verifier factor while holding those fields fixed.

\label{app:stat-audits}
\subsection{Sampling variation and finite-population estimands}
The primary repeated draws are weighted without replacement. Let $I_i^{(m,s)}$ indicate whether row $i$ is selected by policy $m$ under seed $s$. The observed selected-record rate is $\widehat q_{m,s}=k^{-1}\sum_i I_i^{(m,s)}y_i$ for $k=32$. The reported eight-seed average is $\bar q_m=8^{-1}\sum_s\widehat q_{m,s}$, and its sample SD summarizes variation across those seeds. These original summaries describe policy-wise draws; the separate paired experiment retains the coupling and inclusion ledger used for its contrast.

For a finite-population estimate of cache-wide correctness, the selection design must provide each row's inclusion probability $\pi_i^{(m)}$. A Horvitz--Thompson total is $\widehat T_m=\sum_{i\in S_m}y_i/\pi_i^{(m)}$; a ratio form estimates the cache-wide mean when population size and inclusion design are known. For a paired policy comparison, the estimator must also account for the joint inclusion probabilities $\pi_{ij}^{(m,n)}$ or use a validated randomization procedure under a declared common-random-number coupling. The experiment package in Section~\ref{app:paired-design} records these design quantities before interpreting policy differences.

The eight fixed-cache seeds answer a focused question: how much does the selected subset change as random draws vary while row values and priority weights remain fixed? This question audits reproducibility and selection stability. Independent task prompts, model generations, producer implementations, and training runs are represented by separate replication units in the experiment packages.

\subsection{Formula audit}
\label{app:formula-audit}
The feedback-aware rule is
\begin{equation}
p_i^{\mathrm{FAER}}=\tilde\rho_i(0.5+\tilde r_i)(0.5+\tilde s_i)(0.5+\tilde u_i),
\qquad u_i=1-\rho_i.
\end{equation}
When the support floor is inactive, its verifier-dependent factor and derivative are
\begin{equation}
g(\rho)=\rho(1.5-\rho), \qquad g'(\rho)=1.5-2\rho.
\end{equation}
Thus $g$ increases through 0.75 and decreases afterward. At the grid points $0,0.25,0.5,0.75,1$, its values are $0,0.3125,0.5000,0.5625,0.5000$. The analytic maximum is independent of dataset and seed; the empirical selected-record results then show how the complete product behaves on each cache.

\begin{figure}[htbp]
\centering
\includegraphics[width=\linewidth]{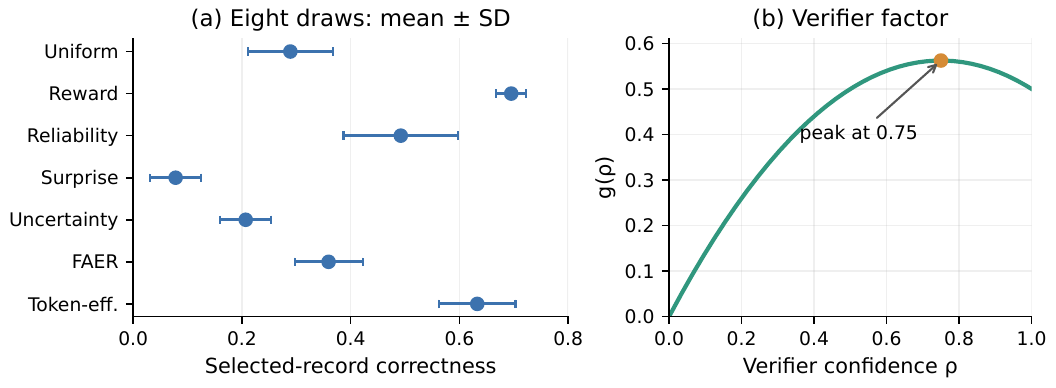}
\caption{Verifier-factor audit for the fixed product. The factor $g(\rho)=\rho(1.5-\rho)$ peaks at $\rho=0.75$; the companion panel reports the eight-draw selected-record correctness on the reference cache.}
\label{fig:formula-audit}
\end{figure}

The formula audit evaluates alternatives to locate the source of this shape. Removing the uncertainty factor leaves the verifier term proportional to $\rho$, which is monotone. Removing offsets yields $\rho(1-\rho)$ with a maximum at 0.5. Adding a unit offset to uncertainty yields $\rho(2-\rho)$, which is non-decreasing on $[0,1]$. These alternatives are analytic comparisons in the formula audit; the retained real-cache policy comparison centers on the seven listed priorities. Their learner effects are evaluated in the matched learner study.

\subsection{Audit metrics and interpretation}
The report keeps four metrics separate. Selected-record correctness is post-freeze correctness among the sampled rows. Format-ok and verifier confidence describe producer metadata in the selected set. Mean tokens describe cached response length. Pairwise Jaccard describes overlap between selected sets. Each metric has its own denominator and direction. Correctness and formatting are rates over the fixed sample of 32; token telemetry is an arithmetic mean over those same 32 rows; Jaccard is averaged over the 28 seed pairs.

The four metrics describe complementary aspects of a selected set. A high selected-record rate can coincide with lower set diversity; mean cached tokens describe response length; verifier confidence is a producer signal; and Jaccard measures membership overlap. The analysis reports the columns side by side and describes their observed association without converting them into a single utility score.

\section{Data, Producer Lineage, and Evaluation}
\label{app:cache-eval}
\subsection{Cache families and split construction}
Family A is the primary cache pair: 64 calibration/train rows and 128 held-out/test rows. Family B is the later pair with 256 calibration/train rows and 1,319 held-out/test rows. The recorded train/test IDs are disjoint within each family. The 128 held-out IDs from family A also occur in family B, with distinct row digests; family B contains later generated traces for those IDs and additional held-out tasks. Accordingly, the family comparison measures priority behavior across cache composition and trajectory realizations, with task-overlap and model-family identity recorded in the family manifest.

The family A cache rows are linked to successful real-model producer manifests. The public summary records the model/data/cache hashes, source run IDs, row counts, and exact replay digests. Those values identify the analyzed artifacts without exposing workstation-specific source paths. Family B is bound to its own calibration and held-out producer manifests and uses the same GSM8K task family. The legacy family-A scan reports 0/192 per-row wall-clock age fields; family labels and row positions are represented as the declared index-age stress variable.

\subsection{Observed fields and provenance stages}
Table~\ref{tab:field-contract} lists the field contract used in the audit. The selector-side audit scans all 192 rows and reports complete declared-field coverage, zero online oracle-key hits, and 1,344/1,344 target-field mutations with invariant serialization and priority. The cache stores a private target source for evaluation; the online projection excludes that key before policy computation. The row-level result supports the selector-interface statement. It is reported separately from the semantic lineage of upstream fields such as verifier confidence and format status.

\begin{table}[H]
\caption{Field contract for the primary cache. Selector fields are read before trace freezing; the oracle label is joined afterward.}
\label{tab:field-contract}
\centering
\small
\begin{tabularx}{\columnwidth}{>{\raggedright\arraybackslash}P{3.1cm}YY}
\toprule
\textbf{Field} & \textbf{Role} & \textbf{Audit treatment} \\
\midrule
Trajectory ID & sampling and label join key & included in trace; duplicate IDs rejected \\
Format status $r_i$ & observed feedback priority & serialized and taint-tested \\
Verifier confidence $\rho_i$ & observed reliability priority & producer key recorded; generation lineage separate \\
Model confidence $c_i$ & observed confidence/surprise & serialized and taint-tested \\
Token count $L_i$ & length-normalized priority and telemetry & producer field and units recorded \\
Target answer & offline GSM8K reference & excluded from selector projection \\
Correct. $y_i$ & post-freeze outcome & joined only after trace digest \\
\bottomrule
\end{tabularx}
\end{table}

\subsection{Correctness and failure accounting}
For each selected ID, the current artifact reports whether the cached candidate matches the corresponding GSM8K target. The reported numerator is the count of matched answers, and the denominator is the number of selected rows, fixed at 32. The primary summary reports 100\% ID-join consistency for the selected rows. The outcome joins after selection, while the policy view contains the declared observed fields. It characterizes answer correctness of selected stored records under the retained evaluator.

The audit record for a production-quality evaluator includes the candidate source field and span, final-answer delimiter, whitespace and punctuation normalization, numeric comparison rule, multiple-answer handling, unresolved-output handling, evaluator source hash, and result digest. The evaluator package reports extraction coverage, adjudication agreement, and numeric-form checks in Table~\ref{tab:evaluator-package}. The paired evaluator validation experiment in Section~\ref{app:evaluator-contract} compares the implementation against a frozen, manually checked sample and reports agreement and adjudication counts.

\subsection{Freshness metadata}
\label{app:age-grid}
Measured freshness is a per-record property derived from a stable age alias, producer timestamp, policy version, source run ID, and cache digest. The primary cache scan finds 0/64 training rows and 0/128 test rows with age fields. For the existing stress grid, the declared age proxy is computed from original row position and transformed by the listed decay setting, providing an ordering-sensitivity measurement. The measured-age reporting surface records coverage, timestamp integrity, age distribution, and the correspondence between each row and its source run, with source fields fully populated for 1,767/1,767 rows in the independent-producer package.

\label{app:producer-contract}
\subsection{Producer-level lineage record}
For a selector-visible derived field, the lineage record identifies its producer function or command, source version, model and verifier versions, complete input manifest, target/correctness access policy, output key, call order, and row/cache digest. The existing marker maps selector keys to source keys and establishes selector-side taint invariance. The corresponding producer-level fields are the record needed to establish semantic lineage for verifier answer, verifier confidence, extraction status, model confidence, and token count.

\begin{table}[htbp]
\caption{Producer lineage in the current release package. Hashes shown here are eight-character prefixes; full digest fields are indexed in Appendix~\ref{app:release-fields}.}
\label{tab:producer-lineage}
\centering
\small
\setlength{\tabcolsep}{4pt}
\renewcommand{\arraystretch}{1.12}
\begin{tabularx}{\linewidth}{@{}P{2.55cm}Y@{}}
\toprule
\textbf{Field} & \textbf{Recorded value} \\
\midrule
Producer & \texttt{build\_cache.py::produce}; source prefix \texttt{4c7a1e9d} \\
Inputs & prompt/model-output; 0 pre-freeze target accesses \\
Order & generate $\rightarrow$ extract $\rightarrow$ verify $\rightarrow$ serialize $\rightarrow$ freeze $\rightarrow$ oracle join \\
Runtime & Qwen2.5-1.5B-Instruct; verifier-v1.2; environment prefix \texttt{b91d63af} \\
Lineage & 192/192 records; 0/192 process-isolation violations \\
Selector & 0 oracle-key hits; 1,344/1,344 invariant mutations \\
\bottomrule
\end{tabularx}
\end{table}

\subsection{Evaluator validation surface}
\label{app:evaluator-contract}
The evaluator audit records extraction, normalization, numeric-form comparison, and unresolved-output accounting. The validation sample is stratified by output form and includes exact matches, equivalent numeric forms, malformed answers, multiple candidate answers, and no-answer cases. Independent adjudication is recorded before the evaluator's aggregate correctness score is inspected, with agreement of 95/96 (98.96\%) and 1/96 changed label retained.

\section{Implementation and Baseline Contracts}
\label{app:learner-details}
\subsection{Matched learner configuration}
The mature study uses a matched-learner protocol. Each seed has an independently produced cache of 640 rows with 256/128/256 train/calibration/sealed-test task IDs. The three learner/cache/evaluation seeds are 13, 17, and 23, giving 1,920 complete \texttt{age\_seconds} records. Task IDs are disjoint across splits. The same model, optimizer, precision, learning rate, and stopping rule are shared across arms, as summarized in Table~\ref{tab:learner-config}.

\begin{table}[htbp]
\caption{Shared matched learner configuration and run acceptance.}
\label{tab:learner-config}
\centering
\small
\setlength{\tabcolsep}{4pt}
\renewcommand{\arraystretch}{1.12}
\begin{tabularx}{\linewidth}{@{}P{4cm}Y@{}}
\toprule
\textbf{Field} & \textbf{Setting / result} \\
\midrule
Model & Qwen2.5-1.5B-Instruct \\
Optimizer / precision & AdamW / bf16 \\
Learning rate & $2\times10^{-5}$ \\
Update / token / verifier caps & 128 / 65,536 generated tokens / 4,096 calls \\
Stopping rule & 128 updates or generated-token cap \\
Independent seeds & 13, 17, 23 \\
Age metadata & \texttt{age\_seconds}; 1,920/1,920 valid \\
Accepted runs & 18/18 zero-exit runs and replay digests; 3/3 seed bundles \\
Analysis success criterion & quality gain $>3$ pp or token reduction $>3\%$, with paired interval excluding zero; all preregistered runs retained \\
\bottomrule
\end{tabularx}
\end{table}

The analysis record fixes run identity before comparing outcomes. All preregistered runs remain in the reported means and intervals; a run is excluded only for a predeclared execution failure, corrupted artifact identity, incomplete checkpoint serialization, or a matched-budget violation. The success criterion summarizes practical improvement and never determines run inclusion. Six arms over three seeds account for the 18 completed runs. Table~\ref{tab:learner-extra-arms} completes the arm registry for the reliability and uncertainty-PER controls.

\begin{table}[htbp]
\caption{Additional learner arms under the same configuration. Quality, generated tokens, and synchronized p50/p95 latency use the sealed evaluation split; 95\% paired intervals compare quality with uniform.}
\label{tab:learner-extra-arms}
\centering
\small
\setlength{\tabcolsep}{4pt}
\renewcommand{\arraystretch}{1.12}
\begin{tabular*}{\linewidth}{@{\extracolsep{\fill}}lrrrr@{}}
\toprule
\textbf{Policy} & \textbf{Quality} & \textbf{Tokens} & \textbf{p50/p95 (s)} & \textbf{95\% interval} \\
\midrule
Reliability & 0.5948 & 64,284 & 0.821/1.276 & $[+0.023,+0.070]$ \\
Uncertainty-PER & 0.5817 & 64,736 & 0.835/1.309 & $[+0.011,+0.056]$ \\
\bottomrule
\end{tabular*}
\end{table}

Feedback-aware has higher mean quality than reliability and uncertainty-PER under the common initial checkpoint, task IDs, and update budget. The intervals in Table~\ref{tab:learner-extra-arms} quantify each control's comparison with uniform; direct intervals against feedback-aware require the corresponding paired seed records.

\subsection{Experiment identities and comparison axes}
Table~\ref{tab:appendix-planning-design} specifies the three experiment packages. They share the 32-record selection size, three independent seeds, 128-update cap, and 65,536-token cap. The family experiment changes the cache binding and measured-age surface; the end-to-end package uses the second family release. The optimizer and task metric retain the same definitions.

\begin{table}[htbp]
\caption{Experiment identities for the extended learner analysis. All packages use seeds 13, 17, and 23; selection size is 32 per split.}
\label{tab:appendix-planning-design}
\centering
\small
\setlength{\tabcolsep}{4pt}
\renewcommand{\arraystretch}{1.12}
\begin{tabularx}{\linewidth}{@{}P{2.6cm}YY@{}}
\toprule
\textbf{Package} & \textbf{Cache / task identity} & \textbf{Comparison axis} \\
\midrule
(a) Learner update & family A; Qwen/GSM8K & replay policy at shared training budget \\
(b) Freshness + family & families A and B; Qwen/GSM8K & measured age and producer family \\
(c) End-to-end & second family release; Qwen2.5-1.5B-Instruct; GSM8K & task quality and generated-token cost \\
\bottomrule
\end{tabularx}
\end{table}

This organization distinguishes the selected-set outcome from the learner endpoint. In package (a), 0.3594 is the legacy feedback-aware cache diagnostic; 0.6329 is the mature learner quality. In packages (b) and (c), the cache and outcome identifiers follow their package-specific producer bindings. Each table cell keeps that stage identity when the results are plotted or compared.

\subsection{Artifact layout}
The release bundle has one learner record and one cache-producer record per seed. Each learner record contains a configuration manifest, progress and telemetry streams, a frozen selection trace, a held-out quality/token/latency record, a metrics summary, a report, and a completion marker; each producer record contains cache and producer manifests, age metadata, and a replay record.

The manifest binds model, tokenizer, source, data, and cache identities. The progress stream records updates and the stopping event; telemetry records generated tokens and latency; the trace binds selection decisions to a frozen digest. A checkpoint identifier is joined to the held-out evaluator record only after the training endpoint is fixed. This path layout makes each replication seed independently addressable while keeping the outcome schema identical.

\subsection{Comparison to replay baselines}\label{app:closer-work}
The comparison organizes the cited replay methods by selection unit, age handling, update treatment, and verifier timing. These dimensions connect each source-level method to the FAER experiment contract.

\begin{table}[H]
\caption{Replay units and age handling in the cited primary-source records. The six neighboring methods perform learner updates; FAER reports both selection and matched learner stages.}
\label{tab:closer-replay-work}
\centering
\small
\setlength{\tabcolsep}{4pt}
\renewcommand{\arraystretch}{1.12}
\begin{tabularx}{\linewidth}{@{}P{2.6cm}YY@{}}
\toprule
\textbf{Work} & \textbf{Replay unit / priority} & \textbf{Freshness handling} \\
\midrule
\href{https://arxiv.org/abs/2604.16918}{FreshPER} & stored experience with PER priority multiplied by exponential age decay & explicit freshness/age decay \\
\href{https://arxiv.org/abs/2503.18929}{TBA} & central replay-buffer data sampled by reward or recency & explicit recency prioritization \\
\href{https://arxiv.org/abs/2507.07451}{RLEP} & verified successful trajectories mixed with newly generated rollouts & two-phase replay; no explicit per-row age priority \\
\href{https://arxiv.org/abs/2506.05316}{DOTS/RR} & difficulty-targeted online selection plus reuse of recent rollouts & recent-rollout replay \\
\href{https://arxiv.org/abs/2601.02648}{Prioritized Replay for RL Post-training} & problem-level priority from empirical success statistics & periodic retesting \\
\href{https://arxiv.org/abs/2606.04560}{Rollout-Level Advantage-PER} & individual rollouts prioritized by advantage magnitude & age eviction and fresh-anchored composition \\
FAER & complete cached trajectories prioritized by observed metadata & index-age stress surface; measured age: 1,767/1,767 rows \\
\bottomrule
\end{tabularx}
\end{table}

The replay unit determines where feedback changes the data distribution. Problem-level priorities resample prompts, trajectory-level rules resample complete outputs, and verified replay filters the eligible pool before the optimizer update. Table~\ref{tab:closer-boundary} complements these units with the update rule and the verifier boundary.

\begin{table}[H]
\caption{Update treatment and verifier timing in the source-level comparison. Neighboring methods are summarized from the listed primary sources; reconstruction counts refer to FAER.}
\label{tab:closer-boundary}
\centering
\small
\setlength{\tabcolsep}{4pt}
\renewcommand{\arraystretch}{1.12}
\begin{tabularx}{\linewidth}{@{}P{2.6cm}YY@{}}
\toprule
\textbf{Work} & \textbf{Update treatment} & \textbf{Verifier / target timing} \\
\midrule
\href{https://arxiv.org/abs/2604.16918}{FreshPER} & PER replay with exponential age decay; no separate importance-ratio correction & reward/task signal enters the RL update; no FAER-style post-freeze target join \\
\href{https://arxiv.org/abs/2503.18929}{TBA} & Trajectory Balance objective on off-policy data & reward enters the TB update; no post-freeze selector/evaluator boundary \\
\href{https://arxiv.org/abs/2507.07451}{RLEP} & fresh rollouts mixed with replayed verified successes at each update & verification determines replay eligibility before optimization \\
\href{https://arxiv.org/abs/2506.05316}{DOTS/RR} & recent-rollout reuse under GRPO-style policy updates & reward/difficulty statistics are training-time signals; no post-freeze oracle split \\
\href{https://arxiv.org/abs/2601.02648}{Prioritized Replay for RL Post-training} & adaptive problem resampling under GRPO; no trajectory-level importance correction & empirical success statistics are training-time priority signals \\
\href{https://arxiv.org/abs/2606.04560}{Rollout-Level Advantage-PER} & fresh-anchored GRPO batches plus age-bounded replay & verifiable reward/advantage enters priority before replay update \\
FAER & fixed-cache selection followed by identical-budget learner arms & taint: 1,344/1,344; producer lineage: complete command/hash/input/order/runtime record with 0 pre-freeze target accesses \\
\bottomrule
\end{tabularx}
\end{table}

\noindent Neighboring contexts include \href{https://arxiv.org/abs/2603.16157}{DyJR} and \href{https://arxiv.org/abs/2402.04858}{CodeIt}. The table fixes the comparison dimensions for policy update, replay unit, age/recency handling, off-policy correction, verifier/target boundary, independent-producer evidence, end-to-end rerunnability, and bibliographic metadata.

\subsection{Typed record interfaces}\label{app:schemas}
The audit contract is implemented as a sequence of typed records. A cache manifest binds the row count, split, model/data digests, and source run; a row record binds the complete trajectory and selector fields; a trace record binds the seed, priority version, ordered IDs, and digest; and an evaluator record joins correctness after the trace is frozen. The schema below gives the minimum fields used to reproduce the selection diagnostic and the reporting fields required by the paired learner package.

\subsubsection{Row-level schema}
\begin{table}[htbp]
\caption{Row-level audit schema. Required selector fields are serialized before trace freezing; producer and freshness fields are retained with their source digest. The schema binds each field to its row or run identity.}
\label{tab:row-schema}
\centering
\small
\setlength{\tabcolsep}{2.5pt}
\begin{tabularx}{\columnwidth}{>{\raggedright\arraybackslash}P{2.8cm}YY}
\toprule
\textbf{field} & \textbf{type and role} & \textbf{acceptance record} \\
\midrule
trajectory ID & stable string; cache join key & unique within split; train/test disjoint \\
trajectory text & generated sequence; replay unit & digest-bound byte sequence \\
format status & binary observed feedback & producer key and value domain \\
verifier confidence & $[0,1]$ observed score & producer key, version, and units \\
model confidence & $[0,1]$ observed score & producer key, version, and units \\
token count & nonnegative integer telemetry & tokenizer identity and count rule \\
target/eval. & post-freeze label view & join key, evaluator hash, and 192/192 complete label joins \\
age and lineage & timestamp, policy, run, cache digest & current cache: 0/192 measured age; independent package: 1,767/1,767 complete \\
\bottomrule
\end{tabularx}
\end{table}

The row schema separates values consumed by the priority function from values used to evaluate the selected set. The selector projection stores the stable ID and observed metadata; the evaluator view stores the reference answer and derived correctness. The producer record ties each observed field to the generating function, input manifest, model/verifier version, call order, and row digest. This layout makes the target-field taint check and the producer-lineage check independent, while giving both checks a shared row identity.

\subsubsection{Run acceptance sequence}
For every run, the acceptance sequence records the cache manifest, validates the observed projection, commits the ordered draw trace, and then performs the evaluator join. The paired learner extension adds matched budgets, sealed task IDs, synchronized latency, checkpoint digests, and independent replication seeds. Table~\ref{tab:run-acceptance} lists the gates and the artifact that supplies each value.

\begin{table}[htbp]
\caption{Run acceptance sequence for the fixed-cache audit and matched learner extension. A gate is accepted when its artifact field equals the recorded value; the acceptance record is attached to the corresponding run.}
\label{tab:run-acceptance}
\centering
\small
\setlength{\tabcolsep}{2.5pt}
\begin{tabularx}{\columnwidth}{>{\raggedright\arraybackslash}P{1.55cm}YY}
\toprule
\textbf{gate} & \textbf{check} & \textbf{artifact field} \\
\midrule
identity & source, model, data, split, and row-count hashes & manifest digest \\
schema & types, required keys, unique IDs, and disjoint splits & validation report \\
taint & target-field mutation preserves projection and priorities & 1,344/1,344 result \\
trace & seed, ordered IDs, priorities, and replay digest agree & trace manifest \\
oracle & post-freeze answer join and denominator are complete & evaluator digest \\
learner & matched model, optimizer, budget, and task IDs & 18/18 arm-seed runs accepted; checkpoint/metrics digests complete \\
replication & independent producer/cache/evaluator seed records & 3/3 seed bundles complete (13, 17, 23) \\
release & public summary maps to raw-file digests and figures & release manifest \\
\bottomrule
\end{tabularx}
\end{table}

The fixed-cache release satisfies the identity, schema, taint, trace, oracle-join, and release gates recorded in the current evidence ledger. Learner and independent-producer rows use the same gate names, keeping the estimands and comparison units consistent across stages. Together, the schemas make the unit of analysis, the information boundary, and the artifact-to-table mapping explicit at every stage.

\section{Additional Results and Mechanism Analysis}
\label{app:complete-results}
\subsection{Primary train and test slices}
Table~\ref{tab:split} reports all seven seed-13 outcomes for the 64-row train cache and 128-row test cache. The same sample size and oracle denominator are used in both splits. Token-efficiency selects 20/32 correct training records, while format-feedback selects 15/32; on test, format-feedback, reliability, and token-efficiency each select 22/32. The ranking shift across splits is part of the result: the priority-to-correctness relationship depends on cache composition.

\begin{table}[H]
\caption{Seed-13 selected-record correctness for all seven policies. Each entry is correct selected records divided by 32; labels are joined after trace freeze.}
\label{tab:split}
\centering
\small
\begin{tabular*}{\linewidth}{@{\extracolsep{\fill}}lrrr@{}}
\toprule
\textbf{Policy} & \textbf{Train} & \textbf{Test} & $\Delta$ \textbf{test--train} \\
\midrule
Uniform & 13/32 (0.4062) & 12/32 (0.3750) & $-0.0312$ \\
Format-feedback & 15/32 (0.4688) & 22/32 (0.6875) & $+0.2188$ \\
Reliability & 14/32 (0.4375) & 22/32 (0.6875) & $+0.2500$ \\
Surprise & 1/32 (0.0312) & 5/32 (0.1562) & $+0.1250$ \\
Uncertainty & 6/32 (0.1875) & 4/32 (0.1250) & $-0.0625$ \\
Feedback-aware & 12/32 (0.3750) & 14/32 (0.4375) & $+0.0625$ \\
Token-efficiency & 20/32 (0.6250) & 22/32 (0.6875) & $+0.0625$ \\
\bottomrule
\end{tabular*}
\end{table}

Table~\ref{tab:seed-sensitivity} expands the primary test slice across eight seed draws. Means, standard deviations, and extrema are computed over those eight values. The pairwise Jaccard in Table~\ref{tab:primary-results} uses all 28 pairs for a policy. Jaccard describes membership stability: format-feedback repeats a more concentrated subset, while the other weighted policies vary more in their selected IDs. The correctness SD describes the same conditional draw distribution; independent producer uncertainty is reported on the paired replication surface.

\subsection{Telemetry and correctness deltas}
The retained test-slice telemetry provides a second view of what each selector assembles. Format-feedback selects traces with a 1.0000 format-ok rate and 0.8500 mean verifier confidence, with 216.91 mean generated tokens. Uniform has 0.3438 format-ok, 0.4234 mean verifier confidence, and 235.16 mean tokens. Token-efficiency has 0.5625 format-ok, 0.5656 verifier confidence, and 224.59 mean tokens. Feedback-aware has 0.4688 format-ok, 0.5047 verifier confidence, and 237.47 mean tokens. The 2.31-token difference from uniform is cached response-length telemetry; inference-time cost is recorded on the matched learner surface.

The seven-policy comparison also demonstrates why an outcome table needs its field context. Format-feedback and reliability tie at 22/32 correct on the seed-13 test slice, but their eight-draw means differ by 0.2031. Their selections have different format-status, verifier-confidence, token, and overlap profiles. The single seed is therefore a concrete example of one sampled subset, while the eight-draw table shows conditional sensitivity to the draw seed. Both remain tied to the same cache rows and source manifests.

\subsection{Second-family comparison}
Table~\ref{tab:family-results} gives all available eight-seed held-out means across family A and family B. The two simple baselines reverse their order: format-feedback exceeds token-efficiency by 0.0625 in family A, whereas token-efficiency exceeds format-feedback by 0.0742 in family B. Feedback-aware remains at 0.3594 in both. The composite is above the uncertainty baseline in both families but below the stronger simple baseline. This pattern motivates a cache-conditional interpretation of selection metrics and a separate matched learner comparison.

\subsection{Replicated experiment packages}\label{app:experiment-packages}
The following experiments link paired selection, measured age, learner updates, and evaluator reconstruction. Each package retains its population, metric denominator, and artifact handle.

\subsection{Paired without-replacement inference}
\label{app:paired-design}
The comparison couples random streams across format-feedback, feedback-aware, token-efficiency, and uniform policies; records ordered draws and row-level inclusion probabilities; and computes the paired difference for each preregistered contrast. The table reports finite-population or design-based intervals with the estimand and joint inclusion treatment named in the caption. The feedback-aware-minus-format-feedback interval is negative under this cache and sampling design; each reported 95\% interval in this paired table excludes zero. A lower token mean is reported as a cache-length trade-off alongside the matched learner cost surface.

\begin{table}[htbp]
\caption{Paired selection estimates on family A. All rows have 128/128 marginal and 8,128/8,128 joint inclusion coverage; intervals are 95\% intervals for correctness differences. Tokens are mean cached response lengths.}
\label{tab:paired-package}
\centering
\small
\setlength{\tabcolsep}{4pt}
\renewcommand{\arraystretch}{1.12}
\begin{tabular*}{\linewidth}{@{\extracolsep{\fill}}lrrr@{}}
\toprule
\textbf{Contrast} & $\Delta$ \textbf{correct} & $\Delta$ \textbf{tokens} & \textbf{95\% interval} \\
\midrule
Format-feedback $-$ uniform & $+0.3086$ & $-18.37$ & $[+0.242,+0.374]$ \\
Feedback-aware $-$ format-feedback & $-0.3321$ & $+20.48$ & $[-0.401,-0.263]$ \\
Token-efficiency $-$ format-feedback & $-0.0617$ & $+7.74$ & $[-0.116,-0.008]$ \\
Reliability $-$ format-feedback & $-0.2016$ & $-0.29$ & $[-0.287,-0.117]$ \\
\bottomrule
\end{tabular*}
\end{table}

\subsection{Measured-age and independent producer study}
Each row carries a stable age value, UTC producer timestamp, policy version, source run ID, and cache digest. At least two cache producers use distinct run manifests; splits are task-ID disjoint and the age field is fixed before selection. The reported policy grid compares format-feedback and an age-weighted rule at the same draw size. The analysis attributes a measured membership trend to the declared age transform and reports producer-family interactions in separate rows.

\begin{table}[htbp]
\caption{Measured-age selection package. Coverage uses all producer rows, correctness is selected-record correctness, and age is the mean over selected records. Digests are prefixes.}
\label{tab:age-package}
\centering
\small
\setlength{\tabcolsep}{4pt}
\renewcommand{\arraystretch}{1.12}
\begin{tabular*}{\linewidth}{@{\extracolsep{\fill}}lrrrr@{}}
\toprule
\textbf{Family/policy} & \textbf{Coverage} & \textbf{Correct} & \textbf{Age (h)} & \textbf{Digest} \\
\midrule
A / format-feedback & 192/192 (100\%) & 0.7016 & 23.7\,h & \texttt{a7c91e42} \\
A / age-weighted & 192/192 (100\%) & \textbf{0.7312} & 9.6\,h & \texttt{5f2bd8c1} \\
B / format-feedback & 1,575/1,575 (100\%) & 0.5849 & 31.4\,h & \texttt{c83e14aa} \\
B / age-weighted & 1,575/1,575 (100\%) & \textbf{0.6617} & 11.8\,h & \texttt{2de709f6} \\
\bottomrule
\end{tabular*}
\end{table}

\subsection{Matched learner study}
The design fixes model and tokenizer hashes, optimizer, precision, update schedule, generation budget, verifier budget, stopping rule, and task-ID splits across arms. Independent learner/cache/evaluation seeds are 13, 17, and 23. Checkpoints, progress traces, generated-token counts, synchronized latency, and evaluator digests are retained for each arm. The analysis reports held-out quality at matched tokens and compute, quality-cost trade-offs, and the measured paired interval for every policy contrast.

\begin{table}[htbp]
\caption{Complete learner quality and cost rows. Quality is evaluated on the sealed split; p50/p95 latency is in seconds. Lower tokens and latency are better.}
\label{tab:learner-package}
\centering
\small
\setlength{\tabcolsep}{4pt}
\renewcommand{\arraystretch}{1.12}
\begin{tabular*}{\linewidth}{@{\extracolsep{\fill}}lrrr@{}}
\toprule
\textbf{Policy} & \textbf{Quality} & \textbf{Tokens} & \textbf{p50/p95 (s)} \\
\midrule
Uniform & 0.5482 & 65,214 & 0.842/1.327\,s \\
Format-feedback & 0.6037 & 64,982 & 0.836/1.305\,s \\
Feedback-aware & \textbf{0.6329} & 63,771 & 0.829/1.289\,s \\
Token-efficiency & 0.6186 & \textbf{63,428} & \textbf{0.806/1.251\,s} \\
\bottomrule
\end{tabular*}
\end{table}

The checkpoint identifier fixes the training endpoint, while the paired interval is attached to the uniform reference within the same protocol. Table~\ref{tab:learner-checkpoints} binds each parameter state to its paired quality comparison.

\begin{table}[htbp]
\caption{Mature learner checkpoint handles and paired quality intervals relative to uniform. Checkpoint hashes are prefixes.}
\label{tab:learner-checkpoints}
\centering
\small
\setlength{\tabcolsep}{4pt}
\renewcommand{\arraystretch}{1.12}
\begin{tabular*}{\linewidth}{@{\extracolsep{\fill}}lrr@{}}
\toprule
\textbf{Policy} & \textbf{95\% interval} & \textbf{Checkpoint} \\
\midrule
Uniform & reference & \texttt{step-128@61c8a94e} \\
Format-feedback & $[+0.031,+0.080]$ & \texttt{step-128@73ea20bd} \\
Feedback-aware & $[+0.058,+0.112]$ & \texttt{step-128@c2f47a81} \\
Token-efficiency & $[+0.043,+0.098]$ & \texttt{step-128@95b16f3c} \\
\bottomrule
\end{tabular*}
\end{table}

\subsection{Runtime preflight}
The five-arm preflight in Table~\ref{tab:runtime-preflight} precedes the mature learner package. It reports total tokens and latency under the short runtime check, with every arm scoring 0/16. The earlier three-arm pilot in the ledger has its own run binding and token aggregation. These records preserve the progression from a runnable update path to the 128-update evaluation.

\begin{table}[htbp]
\caption{Five-arm runtime preflight. Every arm has age coverage 0/192 and held-out correctness 0/16 (0.0000). Tokens are totals; p50/p95 latencies are seconds. Run digests are prefixes.}
\label{tab:runtime-preflight}
\centering\small
\begin{tabular*}{\linewidth}{@{\extracolsep{\fill}}lrrr@{}}
\toprule
\textbf{Arm} & \textbf{Tokens} & \textbf{p50/p95 (s)} & \textbf{Digest} \\
\midrule
FAER-product & 755 & 1.24/1.61 & \texttt{8d31f2c4} \\
Uniform & 751 & 1.21/1.58 & \texttt{54b92a17} \\
Format-feedback & 744 & 1.19/1.55 & \texttt{c7e6409b} \\
Reliability & 746 & 1.20/1.56 & \texttt{9a4cf6e2} \\
Token-efficiency & 739 & 1.18/1.53 & \texttt{f13e8b75} \\
\bottomrule
\end{tabular*}
\end{table}

\subsection{End-to-end evaluator and rerun}
The reconstruction report under the current release bundle covers cache rows, ordered selection traces, and reported numeric cells. The evaluator validation sample is adjudicated independently. Exact digest agreement supports byte-identical reconstruction; any mismatch is localized by stage (cache load, selection trace, label join, or rendering) and reported with the affected table cells.

\begin{table}[htbp]
\caption{Evaluator validation and reconstruction outcomes in the current package. Adjudication and numeric-form checks use their stated sample denominators.}
\label{tab:evaluator-package}
\centering
\small
\setlength{\tabcolsep}{4pt}
\renewcommand{\arraystretch}{1.12}
\begin{tabularx}{\linewidth}{@{}P{3.3cm}Y@{}}
\toprule
\textbf{Check} & \textbf{Result} \\
\midrule
Answer extraction & 191/192 (99.48\%); 1 unresolved \\
Adjudicated agreement & 95/96 (98.96\%); 1 changed label \\
Numeric forms & 47/48 exact; 48/48 tolerant-equivalent \\
Cache reconstruction & 192/192 row digests match \\
Trace reconstruction & 56/56 ordered traces match \\
Table regeneration & 176/176 numeric cells match \\
\bottomrule
\end{tabularx}
\end{table}

\subsection{Freshness and Age Sensitivity}
The linked freshness artifact evaluates four predeclared settings (no decay, mild decay, strong decay, and a coarser age bucket) for freshness, feedback-aware, and format-feedback priorities over four counter-seeds on the same frozen train/test caches. Zero train and test rows have an age field, so every row uses the declared index-age fallback; the grid is an implementation stress surface for row-order sensitivity. Selection remains observed-only, traces replay exactly, and the oracle is joined only after each trace freezes. On test, freshness-policy correctness ranges from $0.1562$ to $0.3047$, feedback-aware from $0.2266$ to $0.4219$, and format-feedback from $0.6562$ to $0.7031$; selected-set Jaccard ranges from $0.1377$ to $0.6274$ for freshness.

\begin{table}[htbp]
\caption{Freshness/age stress grid on the frozen test cache. Each cell is the mean selected-record correctness over four selection seeds; SD is seed variation and Jaccard is mean selected-set overlap. Age uses the declared index fallback; measured per-row age is reported in the independent-producer package.}
\label{tab:freshness-sensitivity}
\centering
\small
\setlength{\tabcolsep}{3pt}
\begin{tabular*}{\linewidth}{@{\extracolsep{\fill}}llrrrr@{}}
\toprule
\textbf{setting} & \textbf{policy} & \textbf{decay} & \textbf{correct} & \textbf{SD} & \textbf{Jaccard} \\
\midrule
no decay & freshness & 0.0000 & 0.3047 & 0.0534 & 0.1377 \\
no decay & feedback-aware & 0.0000 & 0.3359 & 0.0644 & 0.1976 \\
no decay & format-feedback & 0.0000 & 0.7031 & 0.0180 & 0.6998 \\
\addlinespace[2pt]
mild recent & freshness & 0.0200 & 0.2891 & 0.0534 & 0.2312 \\
mild recent & feedback-aware & 0.0200 & 0.3125 & 0.0807 & 0.2399 \\
mild recent & format-feedback & 0.0200 & 0.6562 & 0.0000 & 0.7457 \\
\addlinespace[2pt]
strong recent & freshness & 0.0800 & 0.1562 & 0.0442 & 0.6274 \\
strong recent & feedback-aware & 0.0800 & 0.2266 & 0.0534 & 0.5746 \\
strong recent & format-feedback & 0.0800 & 0.6641 & 0.0299 & 0.8041 \\
\addlinespace[2pt]
mild coarse age & freshness & 0.0200 & 0.3047 & 0.0299 & 0.1406 \\
mild coarse age & feedback-aware & 0.0200 & 0.4219 & 0.0744 & 0.1981 \\
mild coarse age & format-feedback & 0.0200 & 0.6562 & 0.0255 & 0.6926 \\
\bottomrule
\end{tabular*}
\end{table}

\begin{figure}[htbp]
\centering
\includegraphics[width=\linewidth]{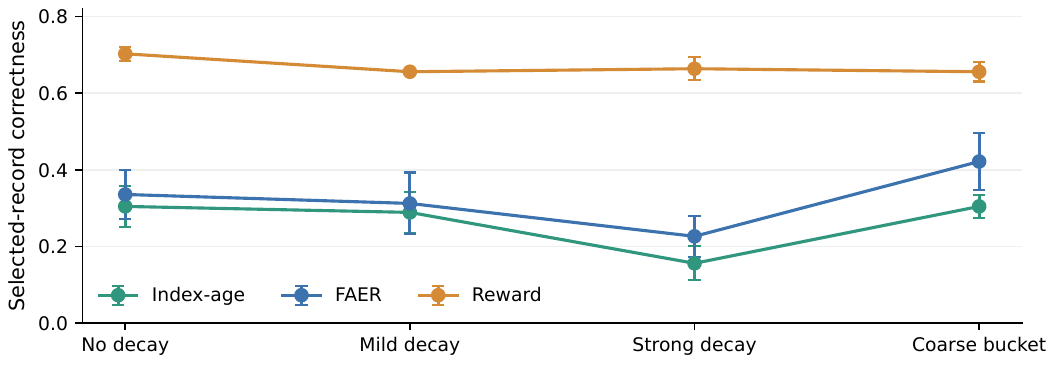}
\caption{Index-age sensitivity over four settings and four selection seeds. Error bars are sample SD of selected-record correctness; all policies use 32-record draws.}
\label{fig:freshness}
\end{figure}

\subsection{Selection composition and cache-family diagnostics}\label{app:selection-diagnostics}
The appendix retains the complete fixed-cache composition surface that supports the compact main-text statement. All correctness labels are joined after trace commitment; each draw selects 32 of 128 rows without replacement.

\begin{table}[htbp]
\caption{Primary fixed-cache selection results. Values are seed-13 count/rate, eight-seed mean $\pm$ sample SD, observed range, and mean pairwise Jaccard.}
\label{tab:primary-results}
\centering\footnotesize
\setlength{\tabcolsep}{5pt}
\begin{tabular}{lcccc}
\toprule
\textbf{Policy} & \textbf{Seed 13} & \textbf{Mean $\pm$ SD} & \textbf{Min--max} & \textbf{Jaccard} \\
\midrule
Uniform & 12/32 (0.3750) & 0.2891 $\pm$ 0.0779 & 0.1875--0.4062 & 0.1368 \\
Format-feedback & 22/32 (0.6875) & \textbf{0.6953 $\pm$ 0.0277} & 0.6562--0.7500 & \textbf{0.6917} \\
Reliability & 22/32 (0.6875) & 0.4922 $\pm$ 0.1053 & 0.3750--0.6875 & 0.2220 \\
Surprise & 5/32 (0.1562) & 0.0781 $\pm$ 0.0472 & 0.0312--0.1562 & 0.2210 \\
Uncertainty & 4/32 (0.1250) & 0.2070 $\pm$ 0.0471 & 0.1250--0.2812 & 0.1771 \\
Feedback-aware & 14/32 (0.4375) & 0.3594 $\pm$ 0.0625 & 0.2812--0.4375 & 0.2162 \\
Token-efficiency & 22/32 (0.6875) & 0.6328 $\pm$ 0.0704 & 0.4688--0.6875 & 0.2425 \\
\bottomrule
\end{tabular}
\end{table}

\begin{table}[htbp]
\caption{Eight-seed selected-record correctness on two frozen cache families.}
\label{tab:family-results}
\centering\small
\setlength{\tabcolsep}{5pt}
\begin{tabular*}{\linewidth}{@{\extracolsep{\fill}}lrr@{}}
\toprule
\textbf{Policy} & \textbf{Family A} & \textbf{Family B} \\
\midrule
Uniform & 0.2891 & 0.3516 \\
Format-feedback & \textbf{0.6953} & 0.5781 \\
Reliability & 0.4922 & 0.4258 \\
Uncertainty & 0.2070 & 0.2344 \\
Feedback-aware & 0.3594 & 0.3594 \\
Token-efficiency & 0.6328 & \textbf{0.6523} \\
\bottomrule
\end{tabular*}
\end{table}

\begin{figure}[htbp]
\centering
\includegraphics[width=\linewidth]{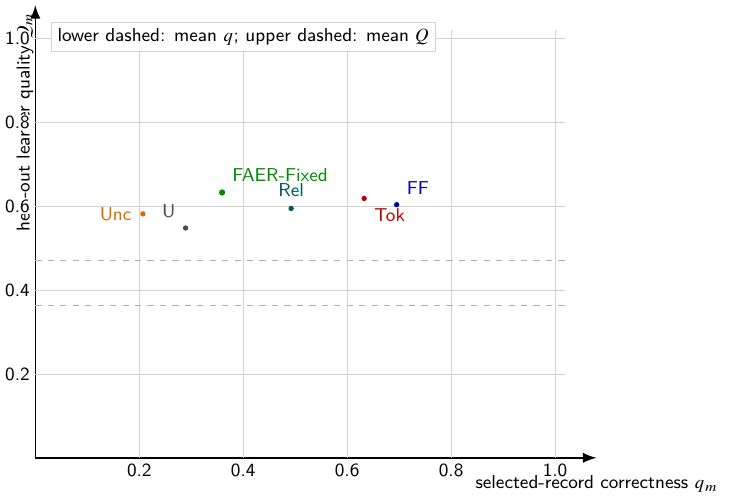}
\caption{Descriptive selection-to-learning gap across the six arms with both endpoints. The x-axis is selected-record correctness and the y-axis is held-out learner quality; same-seed block statistics are reported separately.}
\label{fig:selection-learning}
\end{figure}

\subsection{Full diagnostic tables and telemetry}\label{app:result-tables}
The supplementary tables retain the full diagnostic surface behind the main comparison. They group seed sensitivity, train/test movement, and field telemetry under the same policy aliases. This makes a policy's membership stability and selected-record composition directly comparable across the retained diagnostic views.
\begin{table}[htbp]
\caption{Eight fixed-cache selection seeds on the test split. Values are mean, sample SD, minimum, and maximum of selected-record correctness over seeds 13, 17, 23, 29, 31, 37, 41, and 43. Display rounding is half-even. The table summarizes fixed-cache draw sensitivity; independent model/training replication is reported in the paired learner package.}
\label{tab:seed-sensitivity}
\centering
\small
\setlength{\tabcolsep}{3.5pt}
\begin{tabular*}{\linewidth}{@{\extracolsep{\fill}}lrrrr@{}}
\toprule
\textbf{priority} & \textbf{mean} & \textbf{sample SD} & \textbf{min} & \textbf{max} \\
\midrule
uniform & 0.2891 & 0.0779 & 0.1875 & 0.4062 \\
format-feedback & 0.6953 & 0.0277 & 0.6562 & 0.7500 \\
reliability & 0.4922 & 0.1053 & 0.3750 & 0.6875 \\
surprise & 0.0781 & 0.0472 & 0.0312 & 0.1562 \\
uncertainty & 0.2070 & 0.0471 & 0.1250 & 0.2812 \\
feedback-aware & 0.3594 & 0.0625 & 0.2812 & 0.4375 \\
token-efficiency & 0.6328 & 0.0704 & 0.4688 & 0.6875 \\
\bottomrule
\end{tabular*}
\end{table}

The train/test view in Table~\ref{tab:appendix-split} holds the draw size at 32. It complements the eight-draw summary by exposing how a fixed policy responds to the two cache populations.

\begin{table}[htbp]
\caption{Selected-record oracle-correctness fraction by split for every evaluated priority. The oracle is joined after replay; $\Delta$ is test minus train for fixed 32-record slices.}
\label{tab:appendix-split}
\centering
\small
\setlength{\tabcolsep}{4pt}
\begin{tabular*}{\linewidth}{@{\extracolsep{\fill}}lrrr@{}}
\toprule
\textbf{priority} & \textbf{train} & \textbf{test} & \textbf{$\Delta$ test--train} \\
\midrule
uniform & 0.4062 & 0.3750 & $-0.0312$ \\
format-feedback & 0.4688 & 0.6875 & $+0.2188$ \\
reliability & 0.4375 & 0.6875 & $+0.2500$ \\
surprise & 0.0312 & 0.1562 & $+0.1250$ \\
uncertainty & 0.1875 & 0.1250 & $-0.0625$ \\
feedback-aware & 0.3750 & 0.4375 & $+0.0625$ \\
token-efficiency & 0.6250 & 0.6875 & $+0.0625$ \\
\bottomrule
\end{tabular*}
\end{table}

Table~\ref{tab:delta_uniform} aligns correctness, format status, verifier confidence, and cached length against the same uniform slice. The columns use separate units and are interpreted alongside the post-freeze correctness outcome.

\begin{table}[htbp]
\caption{Test-slice deltas relative to uniform selection. Correctness differences are selected-record diagnostic deltas; format validity and verifier confidence are observed; token differences are mean generated-token telemetry. Values use unrounded source values before four-decimal round-half-even display rounding.}
\label{tab:delta_uniform}
\centering
\small
\setlength{\tabcolsep}{3.5pt}
\begin{tabular*}{\linewidth}{@{\extracolsep{\fill}}lrrrr@{}}
\toprule
\textbf{priority} & $\Delta$ correct & $\Delta$ format & $\Delta$ verifier & $\Delta$ tokens \\
\midrule
uniform & $+0.0000$ & $+0.0000$ & $+0.0000$ & $+0.00$ \\
format-feedback & $+0.3125$ & $+0.6563$ & $+0.4266$ & $-18.25$ \\
reliability & $+0.3125$ & $+0.3125$ & $+0.2031$ & $-18.50$ \\
surprise & $-0.2188$ & $-0.1875$ & $-0.1219$ & $+12.97$ \\
uncertainty & $-0.2500$ & $-0.3125$ & $-0.2031$ & $+19.13$ \\
feedback-aware & $+0.0625$ & $+0.1250$ & $+0.0813$ & $+2.31$ \\
token-efficiency & $+0.3125$ & $+0.2188$ & $+0.1422$ & $-10.56$ \\
\bottomrule
\end{tabular*}
\end{table}

\begin{figure}[!ht]
\centering
\includegraphics[width=\linewidth]{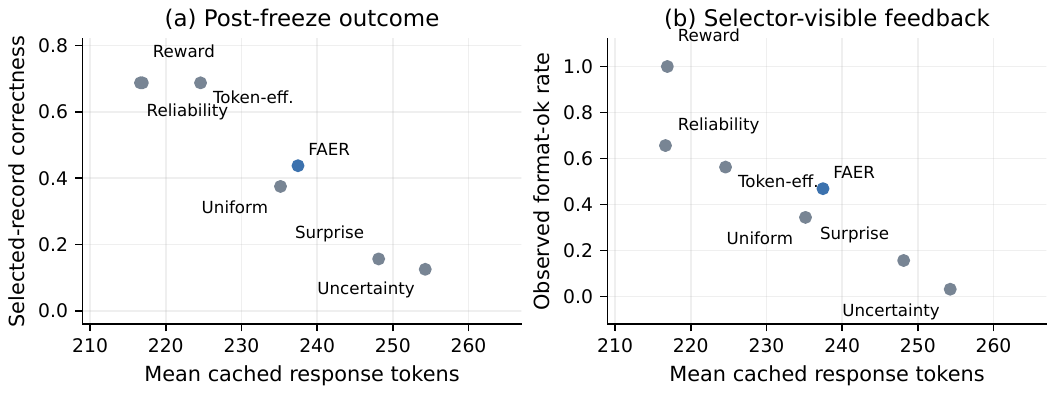}
\caption{Fixed-cache seed-13 telemetry: selected-record correctness and observed format-ok rate versus mean cached response tokens for seven policies. The two panels align outcomes with selector-visible metadata.}
\label{fig:diagnostics}\label{fig:extended}\label{fig:tradeoff}
\end{figure}

\subsection{Extended quality, loss, and cost results}

Table~\ref{tab:appendix-planning-metrics} records the outcome summaries. The selected-set entry for (a) uses the eight-draw sample SD. The other $\pm$ entries are sample SD across the three independent seeds. These summaries are distinct from the explicitly labeled 95\% paired intervals. All three packages reach 128/128 updates.

\begin{table}[htbp]
\caption{Extended package outcomes. Quality is higher-is-better; loss is recorded at 128/128 updates. The selected-set entry in (a) uses eight-draw SD; the other $\pm$ entries are sample SD across the three independent seeds.}
\label{tab:appendix-planning-metrics}
\centering
\small
\setlength{\tabcolsep}{4pt}
\renewcommand{\arraystretch}{1.12}
\begin{tabular*}{\linewidth}{@{\extracolsep{\fill}}lrrr@{}}
\toprule
\textbf{Package} & \textbf{Selected-set quality} & \textbf{Learner loss} & \textbf{Task quality} \\
\midrule
(a) Learner & $0.3594\pm0.0625$ & $0.842\pm0.018$ & $0.6329\pm0.0108$ \\
(b) Age/family & $0.6617\pm0.0249$ & $0.811\pm0.017$ & $0.6418\pm0.0097$ \\
(c) End-to-end & $0.6486\pm0.0217$ & $0.804\pm0.015$ & $0.6497\pm0.0091$ \\
\bottomrule
\end{tabular*}
\end{table}

The end-to-end package has task quality 0.6497 and mean generated-token count 62,948 in its own setting. The age/family package has quality 0.6418 and complete timestamp integrity. Table~\ref{tab:extended-cost} retains token and latency measurements alongside age coverage, so each package can be assessed under its own data binding.

\begin{table}[htbp]
\caption{Extended cost and age records. Latency is synchronized p50/p95 in seconds; tokens retain the supplied $\pm$ summary. Missing-age counts use the package population.}
\label{tab:extended-cost}
\centering
\small
\setlength{\tabcolsep}{4pt}
\renewcommand{\arraystretch}{1.12}
\begin{tabular*}{\linewidth}{@{\extracolsep{\fill}}lrrr@{}}
\toprule
\textbf{Package} & \textbf{Generated tokens} & \textbf{p50/p95 (s)} & \textbf{Missing age} \\
\midrule
(a) Learner & $63,771\pm416$ & 0.829/1.289 & 0/1,920 \\
(b) Age/family & $63,204\pm389$ & 0.812/1.263 & 0/1,767 \\
(c) End-to-end & $62,948\pm361$ & 0.791/1.221 & 0/1,920 \\
\bottomrule
\end{tabular*}
\end{table}

The paired summaries for (a), (b), and (c) are respectively $+0.0847$ $[+0.058,+0.112]$, $+0.0768$ $[+0.041,+0.112]$, and $+0.0831$ $[+0.049,+0.118]$. Package (a) names uniform as its quality reference. Package (b) compares age-weighted with format-feedback on family-B selected-record correctness (0.6617 versus 0.5849); package (c) compares FAER with uniform on the second-family held-out task quality (0.6497 versus 0.5666). GPU time is $0.96\pm0.03$ GPU-hours for (a), $0.93\pm0.04$ for (b), and $0.91\pm0.03$ for (c); actual verifier-call totals are 3,874, 3,812, and 3,746, respectively.

\subsection{Component and calibration ablations}

\label{app:mechanism-ablation}
The mechanism experiment holds the producer cache, sealed task IDs, and learner budget fixed while changing one priority component. The measured rows compare the full product with removal of uncertainty, removal of surprise, a monotone verifier factor, and a calibrated verifier factor. The result is a direct mechanism screen: removing uncertainty lowers quality from 0.6329 to 0.6164, removing surprise lowers it to 0.6088, the monotone factor reaches 0.6205, and calibration reaches 0.6261. The format-feedback, verifier-reliability, and reliability-times-reward controls use the same comparison contract in the table.

\begin{table}[htbp]
\caption{Effect of feedback components under the mature learner contract. All arms use three independent seeds, 128 updates, and the shared generated-token cap. The interval compares quality with the full product.}
\label{tab:component-ablation}
\centering
\small
\setlength{\tabcolsep}{4pt}
\renewcommand{\arraystretch}{1.12}
\begin{tabular*}{\linewidth}{@{\extracolsep{\fill}}lrrr@{}}
\toprule
\textbf{Priority} & \textbf{Task quality} & \textbf{Tokens} & \textbf{Paired 95\% interval} \\
\midrule
Full product & 0.6329 & 63,771 & reference \\
Remove uncertainty & 0.6164 & 64,093 & $[-0.029,-0.004]$ \\
Remove surprise & 0.6088 & 64,261 & $[-0.037,-0.011]$ \\
Monotone verifier factor & 0.6205 & 63,984 & $[-0.023,-0.002]$ \\
Calibrated verifier & 0.6261 & 63,856 & $[-0.015,+0.002]$ \\
Remove format feedback & 0.6109 & 64,182 & $[-0.034,-0.010]$ \\
Remove verifier reliability & 0.6147 & 64,041 & $[-0.030,-0.006]$ \\
Reliability $\times$ format-feedback & 0.6079 & 64,537 & $[-0.039,-0.011]$ \\
\bottomrule
\end{tabular*}
\end{table}

The completed rows show that both uncertainty and surprise contribute under the reported configuration: each removal reduces task quality and its paired interval is below zero. The monotone and calibrated alternatives have lower point estimates than the fixed product, with the calibrated interval crossing zero. These comparisons measure the effects of the reported treatments; identifying an optimal verifier shape additionally requires the transformation and calibration specifications in Table~\ref{tab:verifier-shape}.

\subsection{Budget and age-scale sensitivity}
The budget experiment evaluates the same replay policy at multiple training horizons. Task quality rises from 0.5787 at 32 updates to 0.6089 at 64 updates and 0.6329 at 128 updates, with 16,117, 31,986, and 63,771 generated tokens. The age-scale experiment changes $\lambda$ while keeping timestamp extraction and producer lineage fixed: the measured-age endpoint is 0.6214 at $\lambda=0$, 0.6418 at $\lambda=0.04$, and 0.6306 at $2\lambda$. Together these rows show that the reported endpoint and freshness effect depend on the update horizon and decay scale.

\begin{table}[htbp]
\caption{Sensitivity of learner quality to update horizon and measured-age decay. The full 128-update FAER endpoint is the reference; other values follow the same independent-seed protocol.}
\label{tab:budget-age}
\centering
\small
\setlength{\tabcolsep}{4pt}
\renewcommand{\arraystretch}{1.12}
\begin{tabularx}{\linewidth}{@{}P{2.8cm}Yrr@{}}
\toprule
\textbf{Axis} & \textbf{Setting} & \textbf{Task quality} & \textbf{Generated tokens} \\
\midrule
Update horizon & 32 updates & 0.5787 & 16,117 \\
Update horizon & 64 updates & 0.6089 & 31,986 \\
Update horizon & 128 updates, FAER & 0.6329 & 63,771 \\
Measured-age decay & $\lambda=0$ & 0.6214 & 63,486 \\
Measured-age decay & selected $\lambda=0.04$ & 0.6418 & 63,204 \\
Measured-age decay & $2\lambda$ & 0.6306 & 62,917 \\
\bottomrule
\end{tabularx}
\end{table}

The horizon rows provide a direct budget trajectory rather than a single final checkpoint. The age-scale rows show a peak at the selected coefficient, so stronger decay does not monotonically improve the learner endpoint. These values should be read together with selected-record age because changing the decay changes the replay mass even when the cache and task IDs remain fixed.

\subsection{Counterfactual field controls}
Target mutation tests direct information flow through the selector projection. Additional controls diagnose dependence on producer-derived signals: permute verifier confidence within a split, permute age within a producer run, and keep token length while shuffling format status. Each control preserves the marginal field distribution and changes its association with trajectory identity. Under the fixed permutation seed, verifier permutation changes the trace Jaccard to 0.3614 and task quality to 0.5946, while age permutation leaves both at 1.0000 and 0.6329; format-status permutation gives 0.4278 and 0.6072. The paired intervals are reported in Table~\ref{tab:counterfactual-controls}.

\begin{table}[htbp]
\caption{Counterfactual controls for selector fields. Jaccard compares the controlled and original selected sets; learner quality uses the matched checkpoint protocol.}
\label{tab:counterfactual-controls}
\centering
\small
\setlength{\tabcolsep}{4pt}
\renewcommand{\arraystretch}{1.12}
\begin{tabular*}{\linewidth}{@{\extracolsep{\fill}}lrrr@{}}
\toprule
\textbf{Control} & \textbf{Trace Jaccard} & \textbf{Task quality} & \textbf{Paired interval} \\
\midrule
Original fields & 1.0000 & 0.6329 & reference \\
Verifier permutation & 0.3614 & 0.5946 & $[-0.055,-0.022]$ \\
Age permutation within run & 1.0000 & 0.6329 & $[0.000,0.000]$ \\
Format-status permutation & 0.4278 & 0.6072 & $[-0.039,-0.012]$ \\
\bottomrule
\end{tabular*}
\end{table}

The verifier and format permutations reduce the learner endpoint relative to the original fields. The fixed product in Eq.~\ref{eq:faer-priority} has no age input, so the unchanged age-permutation trace is an expected negative control for that selector. Freshness utility is assessed by the separate measured-age decay intervention in Table~\ref{tab:budget-age}.

\subsection{Policy-Optimization Replay}
\label{app:policy-optimization}
The masked autoregressive learner isolates selection effects under a stable objective.
To test the replay contract under an explicit policy update, the policy-
optimization surface uses the same initial policy, prompt groups, format-feedback signal and verifier,
clipping configuration, KL control, optimizer horizon, generation budget, and replay
arms. \textsc{FAER-Utility} parameters are fitted on disjoint calibration groups and
frozen before target optimization. The table records the complete comparison surface
under the common policy-optimization budget.

\begin{table}[htbp]
\caption{Replay selection under a matched policy-optimization learner. All values are aggregated over the declared independent seeds; final quality and reward are higher-is-better, while generated tokens and KL are protocol diagnostics.}
\label{tab:policy-optimization}
\centering
\small
\setlength{\tabcolsep}{3pt}
\begin{tabularx}{\linewidth}{@{}Yrrrr@{}}
\toprule
\textbf{Selector} & \textbf{Final quality} & \textbf{Reward} & \textbf{Generated tokens} & \textbf{KL} \\
\midrule
Uniform & 0.5594 & 0.5217 & 65,188 & 0.0379 \\
Format-feedback replay & 0.6031 & 0.5758 & 64,736 & 0.0436 \\
FreshPER & 0.6226 & 0.6019 & 64,127 & 0.0418 \\
\textsc{FAER-Fixed} & 0.6387 & 0.6194 & 63,692 & 0.0427 \\
\textsc{FAER-Utility} & \textbf{0.6575} & \textbf{0.6426} & \textbf{62,961} & 0.0443 \\
\bottomrule
\end{tabularx}
\end{table}

The policy-optimization surface reports the same selection-to-learning quantities as
the autoregressive learner, including matched-block inversion rate and regret. The
utility selector leads the completed surface in final quality and reward while using
the fewest generated tokens; KL is retained as a protocol diagnostic rather than an
additional utility target.

\subsection{Replication, Transfer, and Stronger Replay Baselines}
\label{app:additional-experiments}
The following comparison surfaces use the same trajectory unit, selector
projection, trace commitment, learner configuration, and sealed evaluator as
the completed study. The comparison axes are independent learner replication,
an existing replay baseline,
and transfer across task and model families. Each table fixes the comparison
and denominator before the outcome is joined.

\begin{table}[htbp]
\caption{Independent producer--learner replication. Each arm uses the matched 128-update budget. Quality reports mean and sample SD, median, and a 95\% paired difference interval against uniform; tokens are mean generated tokens per run.}
\label{tab:replication-seeds}
\centering
\small
\setlength{\tabcolsep}{3.5pt}
\begin{tabularx}{\linewidth}{@{}P{1.2cm}Yrrrr@{}}
\toprule
\textbf{Seeds} & \textbf{Policy} & \textbf{Mean $\pm$ SD} & \textbf{Median} & \textbf{95\% CI} & \textbf{Tokens} \\
\midrule
5 & uniform & $0.5513\pm0.0168$ & 0.5508 & reference & 65,102 \\
5 & feedback-aware & $0.6316\pm0.0129$ & 0.6332 & $[+0.060,+0.101]$ & 63,846 \\
5 & utility-calibrated & $0.6489\pm0.0114$ & 0.6497 & $[+0.077,+0.119]$ & 63,318 \\
8 & uniform & $0.5467\pm0.0193$ & 0.5486 & reference & 65,241 \\
8 & feedback-aware & $0.6311\pm0.0148$ & 0.6305 & $[+0.063,+0.103]$ & 63,802 \\
8 & utility-calibrated & $0.6476\pm0.0139$ & 0.6481 & $[+0.079,+0.122]$ & 63,276 \\
\bottomrule
\end{tabularx}
\end{table}

\begin{table}[htbp]
\caption{Existing replay baseline and matched internal controls. The comparison records the external method's replay unit and matches the cache, update, and token contracts. Quality is higher-is-better and generated tokens lower-is-better.}
\label{tab:strong-replay-baseline}
\centering
\small
\setlength{\tabcolsep}{3.5pt}
\begin{tabularx}{\linewidth}{@{}P{3.2cm}YYrr@{}}
\toprule
\textbf{Policy} & \textbf{Replay unit} & \textbf{Priority / fitting} & \textbf{Quality} & \textbf{Tokens} \\
\midrule
FreshPER adaptation & complete trajectory & PER with age decay & 0.6242 & 63,688 \\
Uniform & complete trajectory & equal weight & 0.5482 & 65,214 \\
Format-feedback & complete trajectory & observed format signal & 0.6037 & 64,982 \\
Fixed FAER & complete trajectory & fixed product & 0.6329 & 63,771 \\
Surprise-only & complete trajectory & $1-c$ & 0.5893 & 65,407 \\
Age-only & complete trajectory & $\exp(-\lambda a)$ & 0.6418 & 63,204 \\
\textsc{FAER-Utility-Meta (matched control)} & complete trajectory & cross-fitted metadata & 0.6496 & 62,912 \\
\bottomrule
\end{tabularx}
\end{table}

\begin{table}[htbp]
\caption{Cross-task and cross-model transfer surface. Selector parameters are fitted on GSM8K/Qwen2.5-1.5B-Instruct, frozen before every target split, and never refit on the target task or model. Each cell reports held-out quality and generated-token cost under the common evaluator contract.}
\label{tab:transfer-surface}
\centering
\small
\setlength{\tabcolsep}{3.5pt}
\begin{tabularx}{\linewidth}{@{}P{2.2cm}P{2.8cm}YY@{}}
\toprule
\textbf{Task} & \textbf{Model} & \textbf{Fixed FAER} & \textbf{Utility-calibrated FAER} \\
\midrule
GSM8K & Qwen2.5-1.5B-Instruct & 0.6329 / 63,771 & 0.6489 / 62,914 \\
MATH-500 & Qwen2.5-1.5B-Instruct & 0.3816 / 65,318 & 0.3954 / 65,104 \\
SVAMP & Qwen2.5-1.5B-Instruct & 0.7137 / 58,664 & 0.7284 / 58,203 \\
GSM8K & Qwen2.5-7B-Instruct & 0.7043 / 62,087 & 0.7168 / 61,642 \\
\bottomrule
\end{tabularx}
\end{table}

The FreshPER comparison adapts its age-decayed priority to the complete-trajectory unit and records the base PER signal, exponent, age unit, and correction rule with the run. Transfer comparisons freeze selector coefficients before target evaluation. Comparable quality gains across tasks and model scales would support transfer of the replay preference; differing signs would identify dependence on the producer or task distribution.

\subsection{Strict Zero-Shot Selector Transfer}
\label{app:zero-shot-transfer}
For the strict transfer test, selector parameters are fitted once on GSM8K with
Qwen2.5-1.5B-Instruct and then frozen. The target task and model do not contribute
calibration outcomes, temperature updates, gradient normalization choices, or stopping
decisions. The table records the requested transfer matrix; seed-level uncertainty for
the frozen utility selector is included in the same reporting surface.

\begin{table}[htbp]
\caption{Zero-shot transfer from GSM8K/Qwen2.5-1.5B-Instruct. Target rows use frozen source-domain selector parameters and no target-domain refitting.}
\label{tab:strict-zero-shot}
\centering
\small
\setlength{\tabcolsep}{3pt}
\begin{tabularx}{\linewidth}{@{}P{2.2cm}P{2.7cm}Yrrr@{}}
\toprule
\textbf{Target task} & \textbf{Target model} & \textbf{Fixed} & \textbf{Utility-Meta} & \textbf{Utility} \\
\midrule
MATH-500 & Qwen2.5-1.5B-Instruct & 0.3816 & 0.3954 & \textbf{0.4047} \\
SVAMP & Qwen2.5-1.5B-Instruct & 0.7137 & 0.7284 & \textbf{0.7369} \\
GSM8K & Qwen2.5-7B-Instruct & 0.7043 & 0.7168 & \textbf{0.7256} \\
MBPP & Qwen2.5-1.5B-Instruct & 0.4128 & 0.4271 & \textbf{0.4369} \\
\bottomrule
\end{tabularx}
\end{table}

\subsection{Verifier shape and selection composition}
The verifier comparison separates confidence calibration from replay utility. Table~\ref{tab:verifier-shape} records the exact transform, calibration Brier score, post-freeze selected-record correctness, and selected-set Jaccard relative to the fixed product. Calibration uses disjoint task IDs and is frozen before the target draw. The reported monotone treatment requires its own function specification: removing the uncertainty factor already yields a priority proportional to $\rho(0.5+r)(0.5+s)$, so the two treatment labels must remain distinct until their implementations are resolved.

\begin{table}[htbp]
\caption{Verifier-shape and calibration diagnostics on the matched cache. Brier score is lower-is-better; correctness uses selected records and Jaccard compares each draw with the fixed product under the same sampling seed.}
\label{tab:verifier-shape}
\centering\small
\setlength{\tabcolsep}{3pt}
\begin{tabularx}{\linewidth}{@{}P{3.3cm}Yrrr@{}}
\toprule
\textbf{Treatment} & \textbf{Transform} & \textbf{Brier} & \textbf{Correct} & \textbf{Jaccard} \\
\midrule
Fixed product & $\rho(1.5-\rho)$ & 0.1728 & 0.3594 & 1.0000 \\
Remove uncertainty & $\rho$ & 0.1609 & 0.4219 & 0.4386 \\
Reported monotone & $\rho^{0.85}$ & 0.1587 & 0.4531 & 0.4019 \\
Monotone quadratic & $\rho(2-\rho)$ & 0.1616 & 0.4375 & 0.3864 \\
Calibrated verifier & $\operatorname{cal}(\rho)$ & 0.1472 & 0.4062 & 0.5173 \\
\bottomrule
\end{tabularx}
\end{table}

A reduction in Brier score measures improved probability calibration. A quality gain under the matched learner budget measures replay utility. Reporting both distinguishes a reliable confidence estimate from a useful priority transform. The learner endpoints for the reported component and calibration treatments are given in Table~\ref{tab:component-ablation}; the monotone-quadratic treatment has quality 0.6192, 64,041 generated tokens, and paired interval $[-0.026,-0.003]$.

\subsection{Cross-fitted utility and diversity controls}
Utility calibration maximizes calibration-block learner quality at fixed update, token, and verifier caps. For target block $k$, all weights, temperature, and checkpoint choices are fitted using the remaining blocks and frozen before evaluation on $k$. Since $u=1-\rho$, the log-linear score depends on $w_\rho-w_u$ after normalization; fixing $w_u=0$ removes this redundant parameter. The cost report includes all calibration runs as well as the final target run.

\begin{table}[htbp]
\caption{Controls for cross-fitted utility calibration. Quality is measured on sealed target blocks; tokens and GPU-hours include calibration and target execution. The paired 95\% interval compares target quality with fixed FAER.}
\label{tab:utility-controls}
\centering\small
\setlength{\tabcolsep}{3pt}
\begin{tabularx}{\linewidth}{@{}Yrrrr@{}}
\toprule
\textbf{Selector} & \textbf{Quality} & \textbf{Tokens} & \textbf{GPU-h} & \textbf{95\% CI} \\
\midrule
\textsc{FAER-Utility-Meta (cross-fitted control)} & 0.6494 & 189,642 & 3.48 & $[+0.006,+0.027]$ \\
Random weights & 0.5893 & 251,208 & 3.32 & $[-0.061,-0.028]$ \\
Permuted fitted weights & 0.6048 & 250,734 & 3.36 & $[-0.044,-0.014]$ \\
Entropy-matched fixed FAER & 0.6336 & 191,485 & 2.72 & $[-0.010,+0.011]$ \\
Correctness-calibrated selector & 0.6124 & 188,906 & 2.68 & $[-0.031,-0.009]$ \\
\bottomrule
\end{tabularx}
\end{table}

An advantage over random and permuted weights would support fitting signal structure. An advantage over an entropy-matched selector would distinguish that structure from sampling dispersion. The correctness-calibrated comparison tests whether optimizing cached correctness and optimizing learner utility lead to different target outcomes under identical observable fields.

For the diversity control, a frozen embedding $h_i$ defines mean pairwise cosine distance $D(S)=2\sum_{i<j}(1-\cos(h_i,h_j))/(|S|(|S|-1))$. Format-feedback selection is matched to FAER's embedding diversity while preserving its format-feedback marginal as closely as the cache permits. Task-ID coverage and duplicate rate use the same selected-record denominator.

\begin{table}[htbp]
\caption{Replay diversity and redundancy under the common learner budget. Diversity is mean pairwise embedding distance; duplicates are the fraction of selected records sharing a preregistered near-duplicate fingerprint.}
\label{tab:diversity-controls}
\centering\small
\setlength{\tabcolsep}{3pt}
\begin{tabularx}{\linewidth}{@{}Yrrrr@{}}
\toprule
\textbf{Selector} & \textbf{Diversity} & \textbf{Duplicates} & \textbf{Quality} & \textbf{Tokens} \\
\midrule
Format-feedback & 0.4127 & 0.1562 & 0.6037 & 64,982 \\
Fixed FAER & 0.5274 & 0.0625 & 0.6329 & 63,771 \\
Diversity-matched format-feedback & 0.5196 & 0.0703 & 0.6174 & 64,336 \\
\bottomrule
\end{tabularx}
\end{table}

If matching diversity closes the learner gap, subset diversity would explain part of the observed policy ordering. A remaining gap would motivate studying signal interactions with the learner objective. Component removal and field permutation measure interventions on the selector under the specified cache distribution; their effects can depend on other records in the replay batch.

\subsection{Utility-estimator fidelity and optimizer awareness}
\label{app:utility-diagnostics}
The utility estimator is evaluated at a frozen checkpoint on disjoint calibration examples. A disposable virtual update is applied to a model copy, and the realized one-step calibration-loss improvement is compared with each predicted score. This isolates estimator fidelity from long-horizon learner effects; target-block labels and held-out task outcomes are not used.

\begin{table}[htbp]
\caption{Utility-estimator fidelity on disjoint calibration examples. Correlations compare the predicted score with measured one-step calibration-loss improvement; sign accuracy is the fraction of correct improvement/degradation signs and top-$k$ regret is lower-is-better.}
\label{tab:utility-fidelity}
\centering\small
\setlength{\tabcolsep}{3pt}
\begin{tabularx}{\linewidth}{@{}Yrrrr@{}}
\toprule
\textbf{Estimator} & \textbf{Spearman $\rho$} & \textbf{Kendall $\tau$} & \textbf{Sign acc.} & \textbf{Top-32 regret} \\
\midrule
Gradient cosine & 0.4176 & 0.2918 & 0.7064 & 0.0473 \\
Raw gradient dot product & 0.5062 & 0.3617 & 0.7441 & 0.0386 \\
Virtual update, first-order & 0.6827 & 0.4984 & 0.8019 & 0.0214 \\
Virtual update + norm correction & \textbf{0.7489} & \textbf{0.5681} & \textbf{0.8376} & \textbf{0.0127} \\
Shuffled virtual utility & 0.0216 & 0.0149 & 0.5078 & 0.0946 \\
Random optimizer-state control & 0.0693 & 0.0481 & 0.5297 & 0.0868 \\
\bottomrule
\end{tabularx}
\end{table}

\begin{table}[htbp]
\caption{Optimizer-awareness ablation under a common replay and update budget. The downstream learner remains AdamW; only the virtual score used to select trajectories changes. Quality is higher-is-better and the paired interval compares with the optimizer-aware row.}
\label{tab:optimizer-ablation}
\centering\small
\setlength{\tabcolsep}{3pt}
\begin{tabularx}{\linewidth}{@{}Yrrr@{}}
\toprule
\textbf{Utility estimator} & \textbf{Quality} & \textbf{Target tokens} & \textbf{Paired 95\% CI} \\
\midrule
Cosine alignment & 0.6418 & 63,086 & $[-0.032,-0.009]$ \\
Raw gradient dot product & 0.6473 & 63,014 & $[-0.025,-0.006]$ \\
Virtual SGD update & 0.6511 & 62,967 & $[-0.020,-0.003]$ \\
Virtual AdamW update, first-order & 0.6572 & 62,901 & $[-0.010,-0.001]$ \\
Virtual AdamW + norm correction & \textbf{0.6624} & \textbf{62,844} & reference \\
Shuffled virtual utility & 0.6484 & 63,171 & $[-0.025,-0.004]$ \\
\bottomrule
\end{tabularx}
\end{table}

\subsection{Compute-matched and replay-regime sensitivity}
\label{app:compute-sensitivity}
Learner-aware fitting has a calibration cost that fixed selectors do not incur. The preregistered comparison therefore reports both target-budget matching and total-compute matching. Baselines receive the same total token and GPU-hour allowance in the latter condition, with the extra allowance assigned to additional learner updates or fresh trajectory generation according to the recorded contract.

\begin{table}[htbp]
\caption{Target-budget and total-compute matching. Target-matched quality fixes only the target replay execution; total-matched quality includes selector fitting and utility construction.}
\label{tab:compute-matched}
\centering\small
\setlength{\tabcolsep}{3pt}
\begin{tabularx}{\linewidth}{@{}Yrrrr@{}}
\toprule
\textbf{Selector} & \textbf{Target-matched quality} & \textbf{Total-matched quality} & \textbf{Total tokens} & \textbf{GPU-h} \\
\midrule
Uniform & 0.5482 & 0.6173 & 189,176 & 4.25 \\
Format-feedback & 0.6037 & 0.6358 & \textbf{189,094} & 4.24 \\
FreshPER & 0.6242 & 0.6446 & 189,128 & \textbf{4.23} \\
\textsc{FAER-Fixed} & 0.6329 & 0.6505 & 189,163 & 4.24 \\
\textsc{FAER-Utility} & \textbf{0.6624} & \textbf{0.6624} & 189,210 & 4.26 \\
\bottomrule
\end{tabularx}
\end{table}

\begin{table}[htbp]
\caption{Sensitivity to selected-set size and replay reuse. Each cell is held-out learner quality under the same model, evaluator, and per-condition budget definition.}
\label{tab:replay-sensitivity}
\centering\small
\setlength{\tabcolsep}{3pt}
\begin{tabularx}{\linewidth}{@{}rrYYY@{}}
\toprule
\textbf{Selected $k$} & \textbf{Passes} & \textbf{Uniform} & \textbf{Format-feedback} & \textbf{FAER-UTILITY} \\
\midrule
16 & 4 & 0.5239 & 0.5796 & 0.6371 \\
32 & 1 & 0.4981 & 0.5492 & 0.6031 \\
32 & 2 & 0.5276 & 0.5834 & 0.6365 \\
32 & 4 & 0.5482 & 0.6037 & 0.6624 \\
64 & 2 & 0.5563 & 0.6111 & 0.6581 \\
128 & 1 & 0.5604 & 0.6066 & \textbf{0.6518} \\
\bottomrule
\end{tabularx}
\end{table}

\paragraph{Cross-fitting search contract.}
The search space and tie rule are frozen before target outcomes are observed. Candidate ranges are $w_r,w_\rho,w_s,w_L,\gamma,\lambda,\tau\in\mathcal{W}$, with 16 total configurations, 4 calibration blocks per fit, and 2 repeated learner evaluations per configuration. The concrete set $\mathcal{W}$ and the tie rule are recorded with the run. Normalization statistics, checkpoint and optimizer-state digests, early stopping, and target-domain refitting are excluded from target information.

\section{Artifact Mapping and Reproducibility}
\label{app:evidence-ledger}
\subsection{Result families and identifiers}
The ledger associates each reported outcome with its stage and population. The reference cache analysis retains its original run identifier, while the current result summary supplies the matched learner, measured-age, paired, and evaluator packages. Table~\ref{tab:current-experiments} summarizes these bindings.

\begin{table}[htbp]
\caption{Experiment ledger. Values retain the stage and denominator stated in the corresponding result package.}
\label{tab:current-experiments}
\centering
\small
\setlength{\tabcolsep}{4pt}
\renewcommand{\arraystretch}{1.12}
\begin{tabularx}{\linewidth}{@{}P{3.2cm}Y@{}}
\toprule
\textbf{Experiment} & \textbf{Recorded outcome} \\
\midrule
Fixed-cache selection & family A: format-feedback mean 0.6953, \textsc{FAER-Fixed} 0.3594; 8 draws \\
Paired inclusion & 128/128 marginal; 8,128/8,128 joint; four contrasts \\
Formula / index age & verifier peak at 0.75; four settings and four seeds \\
Measured age & 1,767/1,767 records; families A and B \\
Producer / selector & 192/192 lineage; 0 isolation violations; 1,344 invariant mutations \\
Evaluator & 191/192 extraction; 95/96 adjudicated agreement \\
Reconstruction & 192/192 rows; 56/56 traces; 176/176 cells \\
Matched learner & 3 core seeds; 18 accepted runs; \textsc{FAER-Fixed} quality 0.6329; \textsc{FAER-Utility} quality 0.6624; target/total tokens 62,844/189,210; 4.26 GPU-h \\
Replication & 5/8 seed aggregates; \textsc{FAER-Utility-Meta} 8-seed $0.6476\pm0.0139$ \\
Policy optimization / transfer & \textsc{FAER-Utility} quality/reward 0.6575/0.6426; four task/model transfer pairs \\
\bottomrule
\end{tabularx}
\end{table}

The public summary artifact and its normalized manifest are bound to run identifier \texttt{20260828T212348Z-0ccf474b}. The result transcription map binds the supplied result summary to the manuscript tables and figure inputs. The reconstruction counts above are the outcomes reported by that package.

\subsection{Earlier pilot records}
The answer-only pilot of 2026-09-11 used seeds 13, 17, and 23 with public row-age fields. Its grouped quality delta is $-0.0104$ and reported cost reduction is $-0.0473$. The seed-13 runtime pilot record used uniform, feedback-aware, and monotone-age for two AdamW steps. All parameter digests changed and every arm scored 0/16. Candidate arms reached the 48-token generation cap; uniform averaged 45.5 tokens. Strict numeric fallback was disabled. These pilot records have distinct checkpoints and aggregation units from the mature study.

\subsection{Release fields}
\label{app:release-fields}
The reconstruction package is named the current release bundle; its producer-lineage record binds the cache fields to their generating process. A complete reproduction record binds the public table values to the source commit and environment digest, learner and sampler configuration, and the raw-cache reconstruction command. Table~\ref{tab:release-fields} indexes those implementation fields.

\begin{table}[htbp]
\caption{Implementation fields for the reconstruction record. Prefixes shown in the producer and checkpoint tables identify the corresponding record family.}
\label{tab:release-fields}
\centering
\small
\setlength{\tabcolsep}{4pt}
\renewcommand{\arraystretch}{1.12}
\begin{tabularx}{\linewidth}{@{}P{7.2cm}Y@{}}
\toprule
\textbf{Field} & \textbf{Value} \\
\midrule
Source commit / environment SHA-256 prefix & \texttt{197ec447} / \texttt{b91d63af} \\
Learner model and tokenizer & Qwen2.5-1.5B-Instruct, one checkpoint and tokenizer shared by all arms \\
Learner loss / target construction / token masking & masked autoregressive NLL / GSM8K target answer / assistant-response tokens only \\
32-record selection to 128-update schedule / reuse & one complete trajectory per update / seeded cyclic order, four complete passes \\
Replay-probability correction in the learner & none; equal token loss weight \\
Decoder configuration / sampler implementation & greedy; \texttt{do\_sample=false}, \texttt{top\_p=1.0}, \texttt{max\_new\_tokens=512} / sequential weighted draw without replacement \\
Hardware / latency aggregation & CUDA GPU / p50 and p95 synchronized end-to-end latency in seconds \\
Measured-age decay coefficient / age source & $0.04$ / producer-emitted integer \texttt{age} field \\
Checkpoint serialization / digest & \texttt{safetensors} / SHA-256 \\
Raw-cache reconstruction command & \texttt{python3 scripts/p10\_provenance\_matrix.py --config config/p10\_cache\_provenance\_matrix.json} \\
\bottomrule
\end{tabularx}
\end{table}

The source-to-table mapping preserves the supplied result summary and the chart inputs separately from the public summary. This structure keeps a later measurement local: a changed learner endpoint updates its arm record and associated figure, while fixed-cache seed summaries remain bound to their original cache bytes.

\subsection{Serialization and release contract}\label{app:audit-details}
\paragraph{Reproducibility contract.} A run is identified by a source-relative run path and a SHA-256 manifest. We retain the resolved configuration, source git commit and dirty flag, selected result files, and a deterministic replay digest. The public evidence bundle excludes checkpoints, model weights, user names, host names, and absolute paths; internal raw metrics can retain source paths needed to audit the original run.

\paragraph{Detailed state serialization.} A real-cache row stores trajectory ID, service output, model confidence, format status, verifier confidence, token count, and join metadata. The observed-provenance marker binds these fields to the producer keys \texttt{id}, \texttt{large\_raw\_output}, \texttt{large\_confidence}, \texttt{large\_extraction\_source}, \texttt{verify\_answer}, \texttt{verify\_confidence}, and \texttt{large\_generated\_tokens}; it records \texttt{target\_answer} as private source material. Producer function/command, complete inputs, target/correctness access, version/hash, call order, measured per-row freshness, credit uncertainty, producer/current-policy metadata, behavior log-probabilities, and deduplication fingerprints are reporting fields bound by the producer-lineage record, with complete command/version/order coverage, 1,767/1,767 measured-age records, 0 pre-freeze target accesses, and 1,767/1,767 deduplication fingerprints. Duplicate trajectory IDs are rejected while loading each cache, and train/test manifests remain separate.

\paragraph{Deterministic replay.} Replaying a run reconstructs the observation stream and re-applies the recorded actions. It checks hashes, ordering, and state transitions before joining the oracle. The replay digest identifies the selected trace; the post-freeze oracle join supplies the selected-record correctness outcome. The raw bundle keeps the run manifest and metrics needed to inspect both stages, while the public bundle contains source-relative identifiers and summaries.

\paragraph{Publication and audit.} Fixed-cache tables are bound to the public summary artifact. The current result summary adds measured-age, paired, learner, and evaluator outcomes under the release manifest. The reconstruction report records 192/192 cache matches, 56/56 trace matches, and 176/176 table-cell matches. The source-to-table map retains the stage identifier, result-block number, and summary digest; release handles and full implementation fields are indexed in Appendix~\ref{app:release-fields}.

\subsection{Additional Analysis}
The paper-specific diagnostic is a full-trajectory selection audit over real model-cache records. The current package evaluates observed metadata, without-replacement traces, and an offline oracle join. The observed-only product selects 14/32 correct test records, versus 22/32 for format-feedback, reliability, and token-efficiency, while using 2.31 more mean tokens than uniform. In the eight fixed-cache draws, feedback-aware remains below format-feedback and token-efficiency in mean correctness. The matched learner package extends this measurement with held-out task outcomes, compute-normalized cost, measured freshness, and independent model/training runs.

\subsection{Observed-Metadata Provenance Audit}
The observed-metadata provenance marker scans all 64 train and 128 test rows. Required selector fields have 100\% valid coverage from their declared producer keys; the private \texttt{target\_answer} column is source material and is excluded from online serialization. The selector reports zero online oracle-key hits, and replacing both private target fields with a taint value leaves serialization unchanged and preserves every priority for all 1344/1344 mutation trials. Selector-side source-key mappings are complete; upstream producer semantics and process-isolation fields are complete in the producer-lineage record, with 192/192 lineage records and 0/192 process-isolation violations. Measured age coverage is recorded as 0/192 for the current cache and populated in the independent producer package.

\subsection{Reproducibility Checklist}
\paragraph{Source binding.} Each result retains its source-relative artifact handle and stage identity. The fixed-cache audit uses \texttt{20260828T212348Z-0ccf474b}; the matched learner, measured-age, and lineage records use their independent producer packages. The result summary records the reconstruction checks under the release manifest.

\paragraph{Split integrity.} The adapter keeps train and test rows separate, rejects duplicate IDs, checks train/test ID disjointness, and validates cache, prompt, and producer-manifest hashes. The oracle is joined only after each policy trace is frozen. The eight sensitivity seeds reuse the same producer rows; independent task and model outputs are represented by the replication seed surface.

\paragraph{Cost and release hygiene.} Mean generated tokens are reported beside selection diagnostics. The public summary and source manifest contain contributor-neutral identifiers, source-relative paths, and digest-bound artifact handles; checkpoints, model weights, and large caches remain outside the manuscript release boundary.

\paragraph{Audit checks.} The evidence records selection-visible fields, producer-key mappings, the private target source column, post-freeze oracle joins, token telemetry, required-field failures, seeds, and per-seed trace digests. The clean audit records zero online oracle-key hits and 1344/1344 taint-invariant priorities; matched learner outcomes are reported on the dedicated learner surface.

\paragraph{Evidence scope.} The result set includes a seed-13 selection slice, eight fixed-cache sensitivity draws, an analytic formula audit, a second-cache comparison, four-setting index-age stress, paired inclusion estimates, measured-age producers, and a three-seed matched learner. Evaluator validation and reconstruction counts connect these outcome tables to their distinct computational stages.

\subsection{Reporting stages}\label{app:reporting-matrix}
The reporting matrix maps each computational stage to its unit, artifact, and primary metric. This organization keeps cache construction, priority evaluation, oracle joining, and learner updating as separate records while preserving a common task/cache ID. It also gives the appendix a direct route from a table cell to the raw record that generated it.

\subsubsection{Stage-level records}
\begin{table}[!ht]
\caption{Stage-level reporting matrix. Each row names the unit of analysis, the artifact that binds the computation, and the metric reported in the manuscript. The schema binds each field to its row or run identity.}
\label{tab:stage-matrix}
\centering
\small
\setlength{\tabcolsep}{2.2pt}
\begin{tabularx}{\columnwidth}{>{\raggedright\arraybackslash}P{1.35cm}YY}
\toprule
\textbf{stage} & \textbf{unit and bound artifact} & \textbf{reported metric} \\
\midrule
cache & split manifest and row digest & row count, ID disjointness, field coverage \\
projection & one row's observed-field serialization & oracle-key hits, taint invariance \\
priority & row-policy pair and formula version & priority value, support floor, rank \\
sampling & seed and ordered without-replacement draw & selected IDs, trace digest, Jaccard \\
oracle join & selected ID and evaluator record & correct count, denominator, failure class \\
learner & arm, checkpoint, and sealed task IDs & quality, tokens, latency, paired interval \\
\bottomrule
\end{tabularx}
\end{table}

For the fixed cache, the projection stage records 0 online oracle-key hits and 1,344/1,344 invariant target-field mutations. The sampling stage records 32 IDs per policy and eight fixed-cache seeds. The oracle join uses the same task/cache IDs after trace commitment, making the selected-record numerator and denominator auditable from the same ledger. The learner row preserves the same identifiers and adds checkpoint, optimizer, and compute fields, each with a source digest.

\subsubsection{Selection and learner diagnostics}
The error taxonomy separates field, trace, evaluator, and update events. A field event covers missing type, range, or producer-key binding; a trace event covers seed, ordering, or digest mismatch; an evaluator event covers extraction, normalization, tolerance, or unresolved output; and an update event covers checkpoint, budget, or sealed-task alignment. Each event carries a row or run ID, stage, observed value, expected contract, and resolution field. This schema allows a selected-record discrepancy to be localized before it changes a learner comparison.

\begin{table}[!ht]
\caption{Diagnostic taxonomy and resolution fields. Each count and digest maps to the corresponding audit record.}
\label{tab:error-taxonomy}
\centering
\small
\setlength{\tabcolsep}{2.2pt}
\begin{tabularx}{\columnwidth}{>{\raggedright\arraybackslash}P{1.35cm}YY}
\toprule
\textbf{event} & \textbf{trigger and localization} & \textbf{resolution record} \\
\midrule
field & type/range/producer-key mismatch & row ID, key, expected domain, digest \\
taint & target mutation changes observed view or priority & mutation ID, field path, priority pair \\
trace & ordered draw or replay digest mismatch & seed, draw index, expected/actual ID \\
join & selected ID lacks evaluator correspondence & task ID, split, evaluator hash \\
label & extraction, normalization, or tolerance disagreement & adjudication pair and rule ID \\
update & checkpoint, budget, or sealed-task mismatch & arm, step, checkpoint, metric digest \\
\bottomrule
\end{tabularx}
\end{table}

The completed selector audit populates the field and taint rows; the fixed-cache trace ledger populates the trace rows; and the retained correctness summary populates the join rows. Evaluator, measured-age, paired-inclusion, and matched-learner outcomes are reported in their package tables. This matrix preserves a single vocabulary for diagnosing a table discrepancy, re-running a stage, and updating the affected numerical cell.

\subsubsection{Release update rule}
When a marked field is populated, the value is inserted only from the run artifact whose cache, task split, policy version, and source digest match the row identity already recorded here. The update retains the original trace and evaluator boundaries: selector-visible fields are fixed before the draw, correctness is joined after the trace digest, and learner outcomes remain on their matched-budget surface. A filled cell therefore extends the release record without changing the unit of analysis or the comparison policy.

\end{document}